\documentclass{article}

\usepackage{microtype}
\usepackage{graphicx}
\usepackage{booktabs} 

\usepackage{hyperref}

\usepackage[accepted]{icml2024}

\usepackage[export]{adjustbox}
\usepackage{amsmath}
\usepackage{amssymb}
\usepackage{amsthm}
\usepackage{bm}
\usepackage{chngcntr}
\usepackage{enumitem}
\usepackage{dblfloatfix}
\usepackage{mathtools}
\usepackage{multicol}
\usepackage[frozencache=true,cachedir=minted-cache]{minted} 

\usepackage{multirow}
\usepackage{subfig}
\usepackage{tikz}
\usepackage{url}

\makeatletter
\g@addto@macro{\UrlBreaks}{\UrlOrds}
\makeatother

\usepackage[capitalize,noabbrev]{cleveref}

\usetikzlibrary{arrows.meta}
\usetikzlibrary{calc}
\usetikzlibrary{patterns,arrows,decorations.pathreplacing}

\usepackage{afterpage}

\theoremstyle{plain}

\theoremstyle{definition}

\theoremstyle{remark}

\usepackage[textsize=tiny]{todonotes}

\icmltitlerunning{SparQ Attention: Bandwidth-Efficient LLM Inference}

\definecolor{gc_pink}{HTML}{FF6F79}
\definecolor{gc_blue}{HTML}{6FB0FF}
\definecolor{gc_gray}{HTML}{D9D9D9}
\definecolor{gc_dark_pink}{HTML}{99262E}
\definecolor{gc_dark_blue}{HTML}{265A99}
\definecolor{gc_dark_gray}{HTML}{999999}
\definecolor{comment_color}{HTML}{1B8F44}
\definecolor{comment_color_2}{RGB}{64,128,128}

\newcommand{\method}{SparQ Attention}
\newcommand{\methodshort}{SparQ}
\newcommand{\LineComment}[1]{\vspace*{0.85em}\small\textcolor{comment_color_2}{\textit{\# #1}}}
\newcommand{\us}{$\mu$s}
\newcommand{\boldus}{$\boldsymbol{\mu}$\textbf{s}}

\newcommand{\norm}[2]{\lVert #1 \rVert_{#2}}

\makeatletter
\newcommand{\removelatexerror}{\let\@latex@error\@gobble}
\makeatother

\begin{document}

\captionsetup[algorithm]{labelfont=bf}

\twocolumn[
\icmltitle{SparQ Attention: Bandwidth-Efficient LLM Inference}



\icmlsetsymbol{equal}{*}

\begin{icmlauthorlist}
\icmlauthor{Luka Ribar}{equal,graphcore_london}
\icmlauthor{Ivan Chelombiev}{equal,synthesia}
\icmlauthor{Luke Hudlass-Galley}{equal,graphcore_london}
\icmlauthor{Charlie Blake}{graphcore_london}
\icmlauthor{Carlo Luschi}{graphcore_london}
\icmlauthor{Douglas Orr}{graphcore_london}
\end{icmlauthorlist}

\icmlaffiliation{graphcore_london}{Graphcore Research, United Kingdom}
\icmlaffiliation{synthesia}{Synthesia, United Kingdom (work done while at Graphcore Research)}

\icmlcorrespondingauthor{Luka Ribar}{}
\icmlcorrespondingauthor{Luke Hudlass-Galley}{}
\icmlcorrespondingauthor{Douglas Orr}{\space\{lukar, lukehg, douglaso\}@graphcore.ai}

\icmlkeywords{Machine Learning, ICML}

\vskip 0.3in
]



\printAffiliationsAndNotice{\icmlEqualContribution} 

\begin{abstract}
The computational difficulties of large language model (LLM) inference remain a significant obstacle to their widespread deployment. The need for many applications to support long input sequences and process them in large batches typically causes token-generation to be bottlenecked by data transfer. For this reason, we introduce \textbf{\method}, a technique for increasing the inference throughput of LLMs by utilising memory bandwidth more efficiently within the attention layers, through selective fetching of the cached history. Our proposed technique can be applied directly to off-the-shelf LLMs during inference, without requiring any modification to the pre-training setup or additional fine-tuning. We show that \method{} brings up to $8\times$ savings in attention data transfers without substantial drops in accuracy, by evaluating Llama $2$ and $3$, Mistral, Gemma and Pythia models on a wide range of downstream tasks.
\end{abstract}

\begin{figure}[t]
\vspace*{-1.75em}
    \vskip 0.2in
    \begin{center}
        \includegraphics[width=0.95\linewidth]{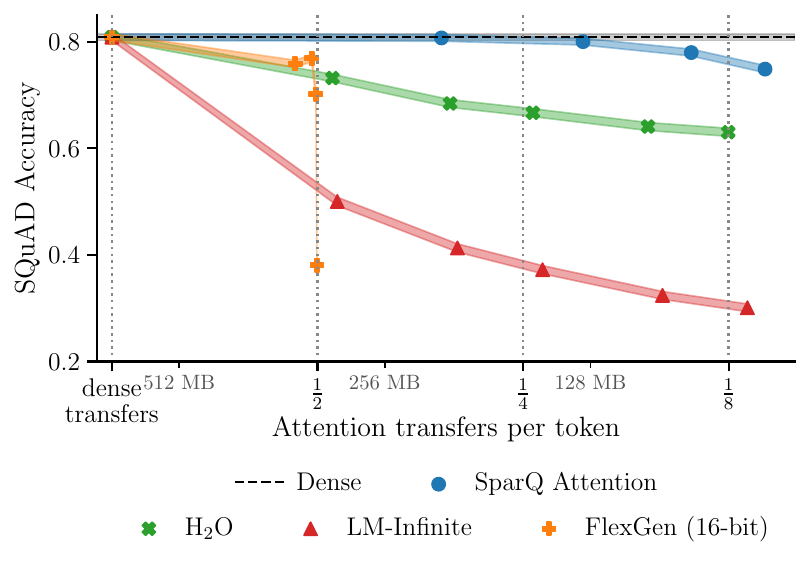}
        \caption{Llama $2$ $13$B SQuAD $1$-shot performance versus attention transfers over a range of compression ratios. \method{} achieves matching performance, while transferring between $1/8$ and $1/4$ as much data as the original dense model. Line thickness shows $\pm$ one standard error over $4000$ examples (the uncertainty from a finite test set). This pattern is representative of the performance across various models and tasks, shown in \cref{fig:app:tradeoff_grid_llama,fig:app:tradeoff_grid_misc,fig:app:tradeoff_grid_pythia}.}
\label{fig:tradeoff_llama_squad}
    \end{center}
    \vskip -0.2in
\end{figure}

\section{Introduction}

Transformer models trained on large corpora of text have recently shown remarkable performance on complex natural language processing tasks \citep{achiam2023gpt, touvron2023llama}. This has been attributed to the \textit{in-context learning} capabilities that emerge with large-scale training, enabling arbitrary textual information (e.g. long instructions, chat histories, relevant documents) to be incorporated at inference-time \citep{wei2022finetuned}.

To leverage the benefits of in-context learning, there has been demand for LLMs to support increasingly long input sequences.
However, the standard inference optimisation used to support in-context learning, \emph{key-value (KV) caching} \citep{pope2023efficientlyscaling}, is constrained by the need to fetch a large amount of data from memory when processing batches of long sequences. This in turn limits the speed at which tokens can be generated---a key usability metric for LLMs.

This bottleneck can be attributed to the auto-regressive nature of transformer generation. For each token generated, the full KV cache must be fetched from memory. The size of the KV cache scales linearly with the sequence length, as well as the batch size, thus rendering generation for long batched sequences increasingly memory bandwidth limited.

Despite this expensive cache-fetch at each step, tokens generally only attend to a small part of the sequence at a time \citep{vig19multiscalevis, yun2020universalsparse}. If it were possible to efficiently predict which tokens will have high attention scores, memory bandwidth efficiency could be significantly increased by only transferring the key-value pairs of high-scoring tokens.

Building upon this idea, we present \textbf{SparQ Attention}, a technique for significantly improving the memory bandwidth efficiency of transformer inference. By approximating attention scores using a subset of query and key components, we fetch only the most relevant tokens for each generation step, reducing the amount of data transferred without degrading the model.

We also provide a new set of challenging downstream task variants which we use to evaluate \method{}. These are based on existing tasks, modified to assess a model's ability to utilise information from long input sequences for multi-token generation. We show that \method{} performs favourably compared to other state-of-the-art methods, giving up to $8\times$ compression without substantial loss in accuracy. \method{} is robust across tasks and models, demonstrated by evaluation on Llama $2$ and $3$, Mistral, Gemma and Pythia. We also provide benchmarks measured on IPU and GPU, showing the practical computational benefits of our approach.

\section{Background}
\label{sec:background}

In this section we provide a straightforward framework to understand the computational efficiency of sequence generation using transformer models (similar to the modelling introduced by \citet{kaplan}) and use it to motivate transfer-efficient attention mechanisms.

\begin{figure}[t]
\begin{center}
    \includegraphics[width=0.9\linewidth]{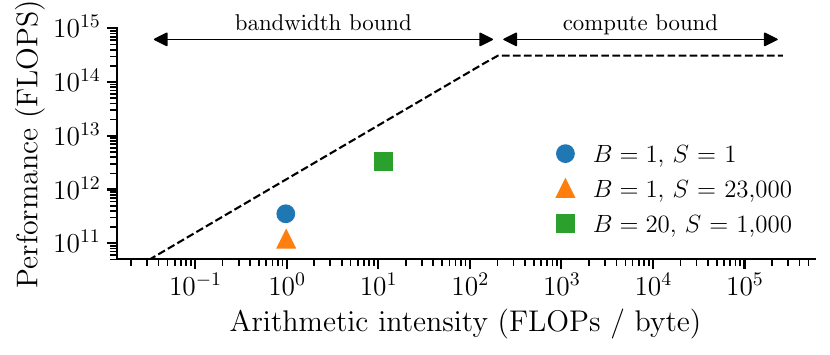}
    \caption{Roofline analysis of Llama $2$ $7$B on A$100$ ($40$GB), highlighting that for a range of LLM inference settings with batch size $B$ and sequence length $S$, practical performance is memory bandwidth bound.}
\label{fig:roofline}
\end{center}
\vspace*{-1em}
\end{figure}

\paragraph{Arithmetic intensity} Consider a compute unit capable of $r_{\mathcal{A}}$ scalar arithmetic operations per second that is connected to a memory via an interface which can transfer $r_{\mathcal{M}}$ scalar elements per second. Given a workload requiring $\mathcal{A}$ arithmetic operations and $\mathcal{M}$ transfers, and assuming concurrent compute and data transfer, the \emph{arithmetic intensity} is defined as $\mathcal{A}/\mathcal{M}$. In LLM inference, $\mathcal{A}$ is primarily a function of the size of matrix multiplications in the model, and $\mathcal{M}$ depends on various factors such as the size of the KV cache, model size, and batch size. When the arithmetic intensity of the workload is less than the ratio $r_{\mathcal{A}}/r_{\mathcal{M}}$, execution time is limited by $r_{\mathcal{M}}$, due to the data transfer taking longer than the computation in the concurrent setting.

The arithmetic intensity of typical sequence generation workloads in transformer models is shown in \cref{fig:roofline}, highlighting that execution time is bandwidth bound, not compute bound. We provide a more general analysis of the arithmetic intensity of sequence generation in \cref{sec:app:arithmetic_intensity}, showing that it is typically bandwidth bound. A corollary of this is that the most effective way to accelerate transformer sequence generation is to reduce data transfers.

\begin{figure}[t]
\begin{center}
    \includegraphics[width=0.9\linewidth]{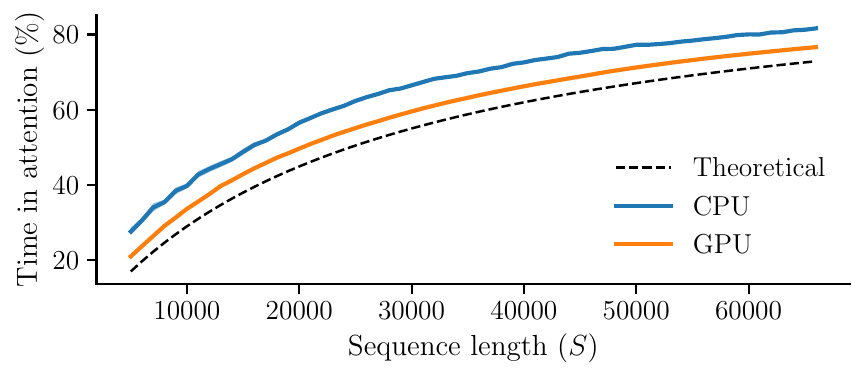}
    \vspace*{-0.25em}
    \caption{The proportion of time that is spent in attention layers during Llama $2$ $7$B inference with a single sample when using \texttt{llama.cpp} on both CPU and GPU platforms. For more details, see \cref{sec:app:llama.cpp}.}
\label{fig:llama_cpp_time_in_attention}
\end{center}
\vspace*{-0.65em}
\end{figure}

\paragraph{Time in attention} Sequence generation with transformers is dominated by two types of computation. The first is a position-wise matrix multiplication between activations and parameters. The second is dot-product self-attention between activations \citep{vaswani2017attention}. Assuming a standard transformer layer with model dimension $d_m$, batch size $B$, sequence length $S$ and using Grouped Query Attention (GQA) \citep{ainslie2023gqa} with $g$ grouped-query heads per key-value head ($g\!=\!1$ for standard multi-head attention), the proportion of data transfers associated with attention is given by
\begin{equation}
    \frac{\mathcal{M}_{\mathrm{attn}}}{\mathcal{M}_{\mathrm{attn}} + \mathcal{M}_{\mathrm{params}}} = \frac{\rho}{\rho + 6/B}\text{,}
\end{equation}
where $\mathcal{M}_{\mathrm{params}}$ and $\mathcal{M}_{\mathrm{attn}}$ are the total data transfers of parameters and KV cache respectively, and $\rho = S / (g d_m)$ is a variable we have introduced to capture the relevant model hyperparameters (see \cref{sec:app:arithmetic_intensity}). When $\rho \gg 6/B$ (for example, with large $S$ or $B$), attention dominates data transfers, as the entire KV cache must be transferred during each generative step. This theoretical trend is backed up by empirical results from \texttt{llama.cpp} \citep{gerganov2024llamacpp} benchmarks in \cref{fig:llama_cpp_time_in_attention}.

Since data transfer is the performance-limiting factor, and attention transfers dominate as sequence length is increased, there is a need for transfer-efficient alternatives to the standard attention mechanism.

\section{Approximating Attention}
\label{sec:approximating_attention}

\begin{figure*}[t]
\centering

\input{approximation_analysis}

\label{fig:approximation_analysis}

\end{figure*}

In this section we examine several properties of the attention operation that enable us to introduce an accurate bandwidth-efficient approximation.

Consider a single attention \emph{query head} with the head dimension $d_h$, processing an input token sequence of length $S$. During autoregressive generation, the output of the attention head is calculated as:
\begin{equation}
    \label{eq:attention}
    \boldsymbol{y} = \mathrm{softmax}\Bigl(\dfrac{\boldsymbol{q}\cdot \boldsymbol{K}^\top}{\sqrt{d_h}}\Bigr) \cdot \boldsymbol{V}
\end{equation}
where $\boldsymbol{q}$ is the query, and $\boldsymbol{K} \in \mathbb{R}^{S \times d_h}$ and $\boldsymbol{V} \in \mathbb{R}^{S \times d_h}$ are the key and value caches respectively. When using GQA \citep{ainslie2023gqa}, $\boldsymbol{K}$ and $\boldsymbol{V}$ are shared across $g$ query heads.

For each forward pass, we need to fetch the key and value matrices from memory, as well as write (append) $\boldsymbol{k}$ and $\boldsymbol{v}$ vectors for the current token, giving a total number of elements transferred per attention head:
\begin{equation}
    \label{eq:memory_dense}
    \mathcal{M}_{\mathrm{dense}} = 2\,S\,d_h + 2\,d_h
\end{equation}
where the first term corresponds to reading the $\boldsymbol{K}$ and $\boldsymbol{V}$ caches and the second term corresponds to writing the current $\boldsymbol{k}$ and $\boldsymbol{v}$ to memory. Memory transfer may be equivalently expressed in terms of bytes, however we use scalar elements to disentangle cache compression methods from the number format used to represent the cache.

\paragraph{Attention scores sparsity}
First, consider the \textit{attention scores} $\boldsymbol{s} \in (0, 1)^{S}$ in \cref{eq:attention}:
\begin{equation}
    \boldsymbol{s} = \mathrm{softmax}\Bigl(\dfrac{\boldsymbol{q}\cdot \boldsymbol{K}^\top}{\sqrt{d_h}}\Bigr)
\end{equation}
Due to the normalising effect of the $\mathrm{softmax}$ function, the resulting $\boldsymbol{s}$ vector is \textit{sparse} (see \cref{fig:approximation_analysis:attention_scores_hist,fig:approximation_analysis:attention_scores_heatmap}), i.e. we can find a boolean mask $\boldsymbol{m_s} \in \{0, 1\}^S$ corresponding to the top-$k$ elements in $\boldsymbol{s}$ ($k \ll S$) such that:
\begin{equation}
    \label{eq:sparse_v}
    \boldsymbol{y}_1 = (\boldsymbol{s} \circ \boldsymbol{m_s}) \cdot \boldsymbol{V} \approx \boldsymbol{s} \cdot \boldsymbol{V}
\end{equation}
As a result, only the values $\boldsymbol{v}_i$ corresponding to the non-zero elements of $\boldsymbol{m_s}$ need to be fetched from memory. However, the algorithm still requires fetching the full $\boldsymbol{K}$ from memory in order to calculate the attention scores $\boldsymbol{s}$, limiting the minimum amount of data transferred to $\frac{1}{2} \mathcal{M}_{\mathrm{dense}}$.

\paragraph{Mean value reallocation}
\begin{table}[t]
    \caption{\emph{Excess} correlation ratio $\eta$ \citep{roche1998correlation} along axes of $\boldsymbol{V}$ (excess: subtract $d^{-0.5}$, so uniform random data $=0.0$). This demonstrates substantial auto-correlation along the sequence axis. Calculated for Llama $2$ $7$B over $40$ SQuAD examples.}
    \label{tab:v_correlation}
    \centering
    \begin{tabular}[b]{cccccc}
    \toprule
     & $B$ & $S$ & Layer & Head & $d_h$ \\\midrule
    $\eta \!-\! d^{-0.5}$ & $0.143$ & $\mathbf{0.256}$ & $0.0$ & $0.0$ & $0.0$ \\\bottomrule
    \end{tabular}
\end{table}
In order to further improve the approximation in \cref{eq:sparse_v}, we note a further observation: $\boldsymbol{v}_i$ vectors within the sequence exhibit a high degree of auto-correlation (see \cref{tab:v_correlation}). Thus, an additional correction term using a running-mean value vector $\boldsymbol{\bar{v}} = \frac{1}{S} \sum_{i=1}^S \boldsymbol{v}_i$ can be added as follows:
\begin{equation}
    \label{eq:sparse_v_reallocation}
    \boldsymbol{y}_2 =  (\boldsymbol{s} \circ \boldsymbol{m_s}) \cdot \boldsymbol{V} + (1 - \boldsymbol{s} \cdot \boldsymbol{m_s}) \bar{\boldsymbol{v}}
\end{equation}
This introduces a minimal additional overhead compared to \cref{eq:sparse_v} due to the mean vector $\bar{\boldsymbol{v}}$ being updated and written back to memory at each step.

\paragraph{Query sparsity}
In order to improve the lower bound on memory transfers, we further consider efficiently approximating the mask $\boldsymbol{m_s}$ by calculating approximate attention scores $\hat{\boldsymbol{s}}$ without using the full matrix $\boldsymbol{K}$. Here, we consider the distribution of magnitudes of the components of the query vector $\boldsymbol{q}$ and observe that it is highly \textit{heavy-tailed} (see \cref{fig:approximation_analysis:query_hist,fig:approximation_analysis:query_kurtosis_strip}). This observation allows us to efficiently approximate the attention scores $\boldsymbol{s}$ by defining a per-query boolean mask $\boldsymbol{m_q} \in \{0, 1\}^{d_h}$ corresponding to the top-$r$ components of $\boldsymbol{q}$. The scores are then approximated as:
\begin{equation}
    \label{eq:sparq_softmax}
    \hat{\boldsymbol{s}} = \mathrm{softmax} \bigg( \dfrac{(\boldsymbol{q} \circ \boldsymbol{m_q}) \cdot \boldsymbol{K}^\top}{\tau} \bigg)
\end{equation}
where $\tau$ is the softmax temperature. Due to the mask $\boldsymbol{m_q}$, only the components of $\boldsymbol{K}$ corresponding to non-zero elements of the mask need to be fetched from memory. The top-k mask $\boldsymbol{m_{\hat{s}}} \in \{0, 1\}^{S}$ can then be calculated using $\hat{\boldsymbol{s}}$ (see \cref{fig:approximation_analysis:agreement_rk_violin}) and the approximate attention output is obtained as:
\begin{equation}
    \label{eq:sparq_no_reallocation}
    \boldsymbol{y}_3 = \mathrm{softmax}\bigg(\dfrac{\boldsymbol{q}\cdot \boldsymbol{K}^\top}{\sqrt{d_h}} + \log(\boldsymbol{m_{\hat{s}}} + \epsilon)\bigg) \cdot \boldsymbol{V}
\end{equation}
with $\epsilon \to 0$.
Again, due to the mask $\boldsymbol{m_{\hat{s}}}$, only the key-value pairs corresponding to the non-masked elements need to be fetched from the memory.
\paragraph{Mean value reallocation with query sparsity}
As a final consideration, we look at combining the mean value reallocation improvement of \cref{eq:sparse_v_reallocation} with the approach in \cref{eq:sparq_no_reallocation}. As we do not have access to the full scores $\boldsymbol{s}$, we proceed to approximate the weighted sum using the approximate scores in \cref{eq:sparq_softmax}. Note that, since the query-key dot product is performed over only $r$ dimensions, care needs to be taken when choosing the appropriate softmax temperature $\tau$ in \cref{eq:sparq_softmax}. If $r$ components were chosen randomly, the appropriate temperature would be $\sqrt{r}$. On the other hand, if the top-$r$ components were the only non-zero elements of the query vector, the appropriate temperature would remain $\sqrt{d_h}$. As a balance between the two extremes, we have found the following temperature to yield a good approximation (see \cref{fig:approximation_analysis:reallocation_scale_scatter}):
\begin{equation}
    \tau = \sqrt{d_h \dfrac{\norm{\boldsymbol{q} \circ \boldsymbol{m_q}}{1}}{\norm{\boldsymbol{q}}{1}}} \label{eq:l1coveragetemp}
\end{equation}
The final attention output can be then calculated as a weighted sum:
\begin{equation}
    \boldsymbol{y} = \alpha\, \boldsymbol{y}_3 + (1 - \alpha) \boldsymbol{\bar{v}}
\end{equation}
where $\alpha = \boldsymbol{m_{\hat{s}}} \cdot \boldsymbol{\hat{s}}$ is the relative weight of the top-$k$ terms.


\section{\method{}}
\label{sec:sparseq}

\begin{figure*}[ht]
    \centering
    \resizebox{0.88\width}{!}{
    \begin{minipage}[t]{\columnwidth}
        \centering
        \raisebox{-\height}{\resizebox{0.9\columnwidth}{!}{

\begin{tikzpicture}

\fill[fill=none] (-2, -4) rectangle (-1, -4) node[pos=0.5] {\LARGE $\bm{q}_{[\textcolor{gc_dark_pink}{\bm{i}_1}]}$};

\filldraw[draw=black, fill=gc_pink] (-0.5, -2) rectangle (0.5, -3) node[pos=0.5] {$\bm{0.8}$};
\filldraw[draw=black, fill=gc_gray] (-0.5, -3) rectangle (0.5, -4) node[pos=0.5] {$-0.2$};
\filldraw[draw=black, fill=gc_pink] (-0.5, -4) rectangle (0.5, -5) node[pos=0.5] {$\bm{-1.3}$};
\filldraw[draw=black, fill=gc_gray] (-0.5, -5) rectangle (0.5, -6) node[pos=0.5] {$0.4$};

\fill[fill=none] (0.5, -4) rectangle (2, -4) node[pos=0.5] {\LARGE $\bm{\otimes}$};

\fill[fill=gc_pink] (2, -2) rectangle (12, -3) node[pos=0.5] {};
\fill[fill=gc_gray] (2, -3) rectangle (12, -4) node[pos=0.5] {};
\fill[fill=gc_pink] (2, -4) rectangle (12, -5) node[pos=0.5] {};
\fill[fill=gc_gray] (2, -5) rectangle (12, -6) node[pos=0.5] {};

\fill[fill=none] (12.5, -4) rectangle (13.75, -4) node[pos=0.5] {\LARGE $\bm{K}_{[ \textcolor{gc_dark_pink}{\bm{i}_1}, :]}$};

\draw[draw=black] (2, -2) rectangle (12, -6);
\draw (3, -2) to (3, -6);
\draw (4, -2) to (4, -6);
\draw (5, -2) to (5, -6);
\draw (6, -2) to (6, -6);
\draw (7, -2) to (7, -6);
\draw (8, -2) to (8, -6);
\draw (9, -2) to (9, -6);
\draw (10, -2) to (10, -6);
\draw (11, -2) to (11, -6);
\draw (2, -3) to (12, -3);
\draw (2, -4) to (12, -4);
\draw (2, -5) to (12, -5);


\fill[fill=none] (2, -1.35) rectangle (12, -1.35) node[pos=0.5] {\Large sequence dimension};
\draw[thick, Triangle-Triangle, color=black] (2, -1.75) -- (12, -1.75);

\draw[thick, -Triangle, color=black] (7, -6.25) -- (7, -7);

\filldraw[draw=black, fill=gc_gray, dashed] (2, -8.75) rectangle (3, -8.35);
\filldraw[draw=black, fill=gc_dark_gray] (3, -8.75) rectangle (4, -7.35);
\filldraw[draw=black, fill=gc_dark_gray] (4, -8.75) rectangle (5, -7.75);
\filldraw[draw=black, fill=gc_gray, dashed] (5, -8.75) rectangle (6, -8.35);
\filldraw[draw=black, fill=gc_gray, dashed] (6, -8.75) rectangle (7, -8.65);
\filldraw[draw=black, fill=gc_dark_gray] (7, -8.75) rectangle (8, -7.55);
\filldraw[draw=black, fill=gc_gray, dashed] (8, -8.75) rectangle (9, -8.35);
\filldraw[draw=black, fill=gc_gray, dashed] (9, -8.75) rectangle (10, -8.55);
\filldraw[draw=black, fill=gc_dark_gray] (10, -8.75) rectangle (11, -7.75);
\filldraw[draw=black, fill=gc_gray, dashed] (11, -8.75) rectangle (12, -8.15);
\draw (1.75, -8.75) to (12.25, -8.75);
\fill[fill=none] (2, -9.2) rectangle (12, -9.2) node[pos=0.5] {\Large approximate attention scores $\bm{\hat{s}}$};

\draw[thick, -Triangle, color=black] (1.75, -8.3) -- (1, -8.3);
\fill[fill=none] (-1, -8.5) rectangle (0, -8.5) node[pos=0.5] {\LARGE $\alpha=\displaystyle\sum_{\mathclap{i \in \textcolor{gc_dark_blue}{\bm{i}_2}}} \hat{s}_i$};

\draw[thick, -Triangle, color=black] (7, -9.65) -- (7, -10.4);

\fill[fill=none] (-2, -12.65) rectangle (-1, -12.65) node[pos=0.5] {\LARGE $\bm{q}$};
\filldraw[draw=black, fill=gc_dark_gray] (-0.5, -10.65) rectangle (0.5, -11.65) node[pos=0.5] {$\bm{0.8}$};
\filldraw[draw=black, fill=gc_dark_gray] (-0.5, -11.65) rectangle (0.5, -12.65) node[pos=0.5] {$\bm{-0.2}$};
\filldraw[draw=black, fill=gc_dark_gray] (-0.5, -12.65) rectangle (0.5, -13.65) node[pos=0.5] {$\bm{-1.3}$};
\filldraw[draw=black, fill=gc_dark_gray] (-0.5, -13.65) rectangle (0.5, -14.65) node[pos=0.5] {$\bm{0.4}$};

\fill[fill=none] (0.5, -12.65) rectangle (2, -12.65) node[pos=0.5] {\LARGE $\bm{\otimes}$};


\fill[fill=gc_gray] (2, -10.65) rectangle (12, -14.65) node[pos=0.5] {};
\fill[fill=gc_blue] (3, -10.65) rectangle (5, -14.65) node[pos=0.5] {};
\fill[fill=gc_blue] (7, -10.65) rectangle (8, -14.65) node[pos=0.5] {};
\fill[fill=gc_blue] (10, -10.65) rectangle (11, -14.65) node[pos=0.5] {};

\fill[fill=none] (12.5, -12.65) rectangle (13.75, -12.65) node[pos=0.5] {\LARGE $\bm{K}_{[:, \textcolor{gc_dark_blue}{\bm{i}_2}]}$};
\draw[draw=black] (2, -10.65) rectangle (12, -14.65);

\draw (3, -10.65) to (3, -14.65);
\draw (4, -10.65) to (4, -14.65);
\draw (5, -10.65) to (5, -14.65);
\draw (6, -10.65) to (6, -14.65);
\draw (7, -10.65) to (7, -14.65);
\draw (8, -10.65) to (8, -14.65);
\draw (9, -10.65) to (9, -14.65);
\draw (10, -10.65) to (10, -14.65);
\draw (11, -10.65) to (11, -14.65);

\draw (2, -11.65) to (12, -11.65);
\draw (2, -12.65) to (12, -12.65);
\draw (2, -13.65) to (12, -13.65);

\draw[thick, -Triangle, color=black] (7, -14.9) -- (7, -15.65);

\filldraw[draw=black, fill=gc_dark_gray] (3, -17.4) rectangle (4, -15.7);
\filldraw[draw=black, fill=gc_dark_gray] (4, -17.4) rectangle (5, -16.7);
\filldraw[draw=black, fill=gc_dark_gray] (7, -17.4) rectangle (8, -16);
\filldraw[draw=black, fill=gc_dark_gray] (10, -17.4) rectangle (11, -16.2);

\draw (1.75, -17.4) to (12.25, -17.4);
\fill[fill=none] (2, -17.85) rectangle (12, -17.85) node[pos=0.5] {\Large sparse attention scores $\bm{s}$};

\fill[fill=none] (2, -18.5) rectangle (12, -18.5) node[pos=0.5] {\LARGE $\bm{\otimes}$};


\fill[fill=none] (12.5, -21.2) rectangle (13.75, -21.2) node[pos=0.5] {\LARGE $\bm{V}_{[:, \textcolor{gc_dark_blue}{\bm{i}_2}]}$};

\fill[fill=gc_gray] (2, -19.2) rectangle (12, -23.2);
\fill[fill=gc_blue] (3, -19.2) rectangle (5, -23.2);
\fill[fill=gc_blue] (7, -19.2) rectangle (8, -23.2);
\fill[fill=gc_blue] (10, -19.2) rectangle (11, -23.2);

\draw[draw=black] (2, -19.2) rectangle (12, -23.2);

\draw (3, -19.2) to (3, -23.2);
\draw (4, -19.2) to (4, -23.2);
\draw (5, -19.2) to (5, -23.2);
\draw (6, -19.2) to (6, -23.2);
\draw (7, -19.2) to (7, -23.2);
\draw (8, -19.2) to (8, -23.2);
\draw (9, -19.2) to (9, -23.2);
\draw (10, -19.2) to (10, -23.2);
\draw (11, -19.2) to (11, -23.2);

\draw (2, -20.2) to (12, -20.2);
\draw (2, -21.2) to (12, -21.2);
\draw (2, -22.2) to (12, -22.2);

\filldraw[draw=black, fill=gc_dark_gray] (-0.5, -19.2) rectangle (0.5, -23.2);
\draw (-0.5, -20.2) to (0.5, -20.2);
\draw (-0.5, -21.2) to (0.5, -21.2);
\draw (-0.5, -22.2) to (0.5, -22.2);

\draw [thick,decoration={brace,mirror,amplitude=10pt},decorate] (-0.85, -19.2) -- (-0.85, -23.2);
\fill[fill=none] (-1.65, -19.2) rectangle (-1.65, -23.2) node[pos=0.5,rotate=90] {\LARGE $\bm{\bar{v}}=\text{mean}(\bm{V})$};

\draw[thick, color=black] (0, -23.2) -- (0, -24.25);
\draw[thick, color=black] (3.5, -23.2) -- (3.5, -24.25);
\draw[thick, color=black] (4.5, -23.2) -- (4.5, -24.25);
\draw[thick, color=black] (7.5, -23.2) -- (7.5, -24.25);
\draw[thick, color=black] (10.5, -23.2) -- (10.5, -24.25);
\fill[fill=none] (2, -24.25) rectangle (2, -24.25) node[pos=0.5] {\LARGE $\bm{\oplus}$};
\draw[thick, -Triangle, color=black] (0, -24.25) -- (1.75, -24.25);
\draw[thick, -Triangle, color=black] (10.5, -24.25) -- (2.25, -24.25);
\draw[thick, -Triangle, color=black] (2, -24.5) -- (2, -25.25);

\fill[fill=none] (2, -25.5) rectangle (2, -25.5) node[pos=0.5] {\Large $\bm{y}$};

\fill[fill=none] (-0.5, -24.7) rectangle (2, -24.7) node[pos=0.5] {\large $\times (1 - \alpha)$};
\fill[fill=none] (3.5, -24.7) rectangle (2.25, -24.7) node[pos=0.5] {\large $\times \alpha$};
\end{tikzpicture}}}
    \end{minipage}
    \hspace{4em}
    \begin{minipage}[t]{0.8\columnwidth}
    \removelatexerror
    \input{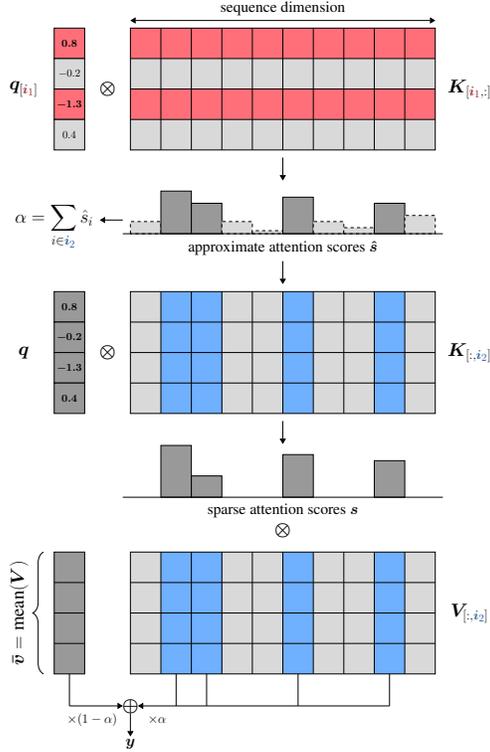}
    \end{minipage}
    }
    \vspace*{-0.8em}
    \caption{\method{} for a single attention head. The algorithm consists of three steps. First, we find the $r$ largest components of the incoming query vector and gather the corresponding components along the hidden dimension of the key cache $\boldsymbol{K}$. This allows us to approximate the full attention scores ($\boldsymbol{\hat{s}}$). In the second step, we identify the top-$k$ largest scores in the approximation and proceed to gather the corresponding full key and value vectors from the cache. As a final step, to compensate for the missing value vectors, we additionally maintain and fetch the running mean value vector $\bm{\bar{v}}$ and reassign it the leftover mass based on approximate score weightings. The attention output is then calculated as usual using the top-$k$ fetched key and value pairs, together with $\bm{\bar{v}}$.}
    \label{fig:schematic}
    \vspace*{-1em}
\end{figure*}

Following the analysis in \cref{sec:approximating_attention}, we propose \textbf{\method{}} (see \cref{fig:schematic}) consisting of three steps:

\begin{enumerate}[label=\textbf{Step \arabic*:},leftmargin=*]
    \item Find the indices of $r$ largest components of $|\bm{q}|$\footnote{We use $|\bm{\cdot}|$ to denote element-wise absolute value.} and only fetch $\boldsymbol{K}$ along the dimensions corresponding to these indices. Calculate \emph{approximate} attention scores $\boldsymbol{\hat{s}}$ using the sliced query and keys.
    \item Find the top-$k$ positions in the approximate attention scores and fetch the corresponding \emph{full} key and value vectors. Calculate the output of the attention operation using the top-$k$ keys and values.
    \item Estimate the total score $\alpha$ assigned to the top-$k$ positions using the approximate attention scores. Use this total score to interpolate between the attention output from the top-$k$ positions, and a \emph{mean value} vector, $\overline{\boldsymbol{v}}$.
\end{enumerate}

The memory transfer of the \method{} algorithm for a single attention head forward-pass:
\begin{equation}
    \mathcal{M}_{\mathrm{\methodshort{}}} = S\,r\, + 2\,k\,d_h + 4\,d_h
\end{equation}
where the first term corresponds to reading $r$ rows of $\boldsymbol{K}$, the second term corresponds to reading the top-$k$ columns of $\boldsymbol{K}$ and $\boldsymbol{V}$ and the third term corresponds to transfers associated with writing the current $\boldsymbol{k}$ and $\boldsymbol{v}$, in addition to reading and writing $\overline{\boldsymbol{v}}$.

By varying $r$ and $k$, we can tune the total amount of data transferred by the scheme, trading-off approximation accuracy for token-generation speed-up. Since typically $S \gg d_h$, $r$ is the most important parameter controlling the data transfer compression ratio $\mathcal{M}_{\mathrm{\methodshort{}}} / \mathcal{M}_{\mathrm{dense}}$.

\paragraph{Grouped query attention}
For models using GQA, groups of $g$ queries access the same KV head. In order to accommodate this, we modify \textbf{Step 1} to sum $|\boldsymbol{q}|$ within each group before selecting top-$r$ components. Similarly, \textbf{Step 2} is modified by summing the approximate attention scores within each group before selecting top-$k$ keys and values for each KV head. Although \textbf{Step 3} can be implemented exactly as before, we found that GQA models obtained better performance without it, so we omitted this step for Llama $3$ and Mistral.
The full code can be found in \cref{sec:app:code}.

\section{Experiments}
\label{sec:experiments}

\begin{table*}
\caption{Results for the largest model of each family tested are presented below. SQuAD and TriviaQA measure performance in accuracy as a percentage; CNN/DailyMail uses ROUGE-L score; WikiText task measures perplexity in bits per character (BPC); Repetition counts the number of characters before the generation diverges. Values presented are those closest to the target compression ratio for each technique, where bold represents the best score for each setting. Median standard errors across all models and sparsity settings are: SQuAD $0.8$, TriviaQA $0.8$, CNN/DailyMail $0.4$, WikiText $0.01$, Repetition $2$.}
\label{table:results}
\resizebox{\linewidth}{!}{%
{\renewcommand{\arraystretch}{0.97}
\begin{tabular}{ crccccccccccccccc }
\toprule
\multicolumn{2}{r}{\textbf{Dataset Name}} & \multicolumn{3}{c}{SQuAD $\uparrow$} & \multicolumn{3}{c}{TriviaQA $\uparrow$} & \multicolumn{3}{c}{CNN/DailyMail $\uparrow$} & \multicolumn{3}{c}{WikiText $\downarrow$} & \multicolumn{3}{c}{Repetition $\uparrow$}\\
\cmidrule(lr){1-2}
\cmidrule(lr){3-5}
\cmidrule(lr){6-8}
\cmidrule(lr){9-11}
\cmidrule(lr){12-14}
\cmidrule(lr){15-17}
\multicolumn{2}{r}{\textbf{Compression}} & $1$ & $1/2$ & $1/8$ & $1$ & $1/2$ & $1/8$ & $1$ & $1/2$ & $1/8$ & $1$ & $1/2$ & $1/8$ & $1$ & $1/2$ & $1/8$\\\midrule
\multirow{2}*{Llama $2$} & LM-$\infty$ & $\bm{80.8}$ & $50.0$ & $30.1$ & $\bm{78.7}$ & $73.4$ & $68.1$ & $\bm{22.1}$ & $16.8$ & $14.9$ & $\bm{0.61}$ & $0.64$ & $0.71$ & $\bm{229}$ & $76$ & $29$\\
\multirow{2}*{$13$B} & H$_2$O & $\bm{80.8}$ & $73.2$ & $63.0$ & $\bm{78.7}$ & $78.5$ & $\bm{78.4}$ & $\bm{22.1}$ & $22.2$ & $20.3$ & $\bm{0.61}$ & $\bm{0.61}$ & $\bm{0.64}$ & $\bm{229}$ & $61$ & $26$\\
& \textbf{\methodshort{}} & $\bm{80.8}$ & $\bm{80.7}$ & $\bm{74.9}$ & $\bm{78.7}$ &  $\bm{78.8}$ & $78.2$ & $\bm{22.1}$ & $\bm{22.5}$ & $\bm{21.6}$ & $\bm{0.61}$ & $\bm{0.61}$ & $0.70$ & $\bm{229}$ & $\bm{227}$ & $\bm{190}$\\\midrule
\multirow{2}*{Llama $3$} & LM-$\infty$ & $\bm{81.2}$ & $66.0$ & $51.8$ & $\bm{83.2}$ & $81.8$ & $80.6$ & $\bm{23.4}$ & $17.1$ & $16.1$ & $\bm{0.56}$ & $0.58$ & $0.64$ & $\bm{213}$ & $102$ & $27$\\
\multirow{2}*{$8$B} & H$_2$O & $\bm{81.2}$ & $74.5$ & $61.7$ & $\bm{83.2}$ & $\bm{83.2}$ & $82.5$ & $\bm{23.4}$ & $23.3$ & $21.9$ & $\bm{0.56}$ & $\bm{0.56}$ & $0.59$ & $\bm{213}$ & $67$ & $30$\\
& \textbf{\methodshort{}} & $\bm{81.2}$ & $\bm{81.2}$ & $\bm{78.3}$ & $\bm{83.2}$ & $83.0$ & $\bm{82.8}$ & $\bm{23.4}$ & $\bm{23.4}$ & $\bm{23.4}$ & $\bm{0.56}$ & $0.57$ & $\bm{0.58}$ & $\bm{213}$ & $\bm{214}$ & $\bm{213}$\\\midrule
\multirow{2}*{Mistral} & LM-$\infty$ & $\bm{81.0}$ & $51.0$ & $29.0$ & $\bm{80.9}$ & $75.8$ & $72.6$ & $\bm{23.7}$ & $18.0$ & $16.6$ & $\bm{0.62}$ & $0.65$ & $0.72$ & $\bm{231}$ & $81$ & $20$\\
\multirow{2}*{$7$B} & H$_2$O & $\bm{81.0}$ & $71.2$ & $56.9$ & $\bm{80.9}$ & $\bm{80.8}$ & $\bm{80.2}$ & $\bm{23.7}$ & $\bm{23.5}$ & $22.8$ & $\bm{0.62}$ & $\bm{0.63}$ & $0.66$ & $\bm{231}$ & $38$ & $14$\\
& \textbf{\methodshort{}} & $\bm{81.0}$ & $\bm{80.9}$ & $\bm{77.5}$ & $\bm{80.9}$ &  $\bm{80.8}$ & $79.0$ & $\bm{23.7}$ & $\bm{23.5}$ & $\bm{23.0}$ & $\bm{0.62}$ & $\bm{0.63}$ & $\bm{0.65}$ & $\bm{231}$ & $\bm{209}$ & $\bm{201}$\\\midrule
\multirow{2}*{Gemma} & LM-$\infty$ & $\bm{80.4}$ & $64.2$ & $48.7$  & $\bm{82.8}$ & $81.6$ & $80.8$ & $\bm{17.4}$ & $13.1$ & $13.3$ & $\bm{0.59}$ & $0.61$ & $0.68$ & $\bm{245}$ & $101$ & $23$\\
\multirow{2}*{$7$B} & H$_2$O & $\bm{80.4}$ & $73.7$ & $60.2$ & $\bm{82.8}$ & $\bm{82.9}$ & $82.5$ & $\bm{17.4}$ & $17.4$ & $16.9$ & $\bm{0.59}$ &  $\bm{0.59}$ & $0.62$ & $\bm{245}$ & $68$ & $18$\\
& \textbf{\methodshort{}} & $\bm{80.4}$ & $\bm{80.3}$ & $\bm{80.3}$ & $\bm{82.8}$ & $82.8$ & $\bm{82.7}$ & $\bm{17.4}$ & $\bm{17.9}$ & $\bm{18.0}$ & $\bm{0.59}$ & $\bm{0.59}$ & $\bm{0.59}$ & $\bm{245}$ & $\bm{240}$ & $\bm{237}$\\\midrule
\multirow{2}*{Pythia} & LM-$\infty$ & $\bm{57.8}$ & $38.5$ & $17.0$ & $\bm{52.6}$ & $41.6$ & $29.7$ & $\bm{20.2}$ & $14.9$ & $14.0$ & $\bm{0.68}$ & $0.71$ & $0.79$ & $\bm{150}$ & $64$ & $18$ \\ 
\multirow{2}*{$6.9$B} & H$_2$O & $\bm{57.8}$ & $52.9$ & $45.5$ & $\bm{52.6}$ & $\bm{52.6}$ & $\bm{52.3}$ & $\bm{20.2}$ & $20.3$ & $18.5$ & $\bm{0.68}$ & $0.69$ & $0.71$ & $\bm{150}$ & $47$ & $17$\\
& \textbf{\methodshort{}} & $\bm{57.8}$ & $\bm{58.0}$ & $\bm{57.1}$ & $\bm{52.6}$ & $52.4$ & $51.7$ & $\bm{20.2}$ & $\bm{20.6}$ & $\bm{20.6}$ & $\bm{0.68}$ & $\bm{0.68}$ & $\bm{0.70}$ & $\bm{150}$ & $\bm{151}$ & $\bm{144}$\\\midrule
\end{tabular}
}
}
\end{table*}

\subsection{Setup}
\paragraph{Models}
We evaluate our method on five widely-used open-source language model variants: Llama $2$ \citep{touvron2023llama}, Llama $3$ \citep{llama3blog}, Mistral \citep{jiang2023mistral}, Gemma \citep{mesnard2024gemma} and Pythia \citep{biderman2023pythia}, evaluating model sizes up to $13$ billion parameters.\footnote{\url{https://github.com/graphcore-research/llm-inference-research/tree/2024-05-sparq}} All models are decoder-only transformers \citep{gpt1}, pre-trained on causal language modelling. They share similar architectural components such as rotary positional embedding \citep{rope}, while also having some notable differences such as different attention mechanisms (multi-head and grouped query attention), layer normalisation implementations, activation functions and execution of modules in parallel.

\paragraph{Tasks}
In order to evaluate our method on a spectrum of relevant NLP tasks that present a particular challenge to sparse attention techniques, our evaluation setup consists of various tasks requiring information retrieval and reasoning over long input sequences. This includes question answering, summarisation, perplexity/bits-per-character (BPC), and text repetition. For this, we adapted standard downstream tasks and datasets to generate examples of sequence lengths between $1$k and $2$k tokens. To define the tasks independently of the selected models, our examples were chosen to have sequence lengths between $4000$ and $8000$ characters, roughly giving the desired lengths in tokens.

For question answering, we use the SQuAD \citep{rajpurkar2016squad} and TriviaQA \citep{joshi2017triviaqa} datasets in the \textit{open-book} setting. In order to construct the SQuAD examples, we augment the provided context (i.e. the standard SQuAD input sequence required to answer the question) with seven additional ``\emph{confusion contexts}'' from unrelated questions. This ensures that the examples have a large sequence length, while making the task harder as the model needs to distinguish the relevant information from the context from the unrelated paragraphs. We use SQuAD v$1.1$, as it does not include unanswerable questions included in SQuAD v$2.0$, since we aim to measure the model's ability to extract useful information from the KV cache. For both question answering tasks we use exact string match accuracy as the evaluation metric. Summarisation is evaluated on the CNN/DailyMail dataset \citep{see2017get} using the ROUGE-L F-score \citep{lin2004rouge} as the metric. We use the WikiText-$103$ dataset \citep{merity2016pointer} with bits per character (BPC) for evaluating language modelling performance.\footnote{We quote performance for sub-word language modelling in BPC, to account for any differences in vocabulary across models.} Finally, we construct an artificial ``Text Repetition'' task to evaluate the capability of the model to repeat sentences from its context verbatim. Such a task can commonly appear in a dialogue setting where the LLM agent is required to retrieve a piece of text from a possibly long context provided, and can be challenging for sparse attention techniques. We construct examples using the Tiny-Shakespeare dataset \citep{karpathy2015unreasonable} by chunking the text into contexts of the appropriate size, appending them with the prompts containing a subset of the context, and evaluating the output exact character length match with the continuation from the context.

\paragraph{Baselines}
We consider the cache eviction technique H$_2$O \citep{zhang2023h2o}, top-$k$ sparse attention in the form of FlexGen \citep{sheng2023flexgen}, and LM-Infinite, a local windowing scheme with initial-tokens included proposed by \citet{han2023lminfinite} as baselines. For each experiment we fix the KV cache transfer budget $k$ idependently of the sequence length. With H$_2$O, we set the local window size $l=k/4$ (with $3k/4$ heavy hitters), and for LM-Infinite we always include the first $16$ positions (with $k-16$ local positions). Due to the lower bound of FlexGen's compression ratio being $1/2$, we do not report the technique's results in \cref{table:results} and \cref{table:needle-in-a-haystack}, but full results can be found in \cref{sec:app:results}. The compression ratio definitions for each of these techniques can be found in \cref{sec:app:method}.

\subsection{Results}
\label{sec:results_main_body}
 
Our experiments span eight distinct models: Llama $2$ with $7$ and $13$ billion parameters, Llama $3$ with $8$ billion parameters, Mistral with $7$ billion parameters, Gemma with $7$ billion parameters, and three Pythia models with $1.4$, $2.8$ and $6.9$ billion parameters. Results from the largest models are presented in \cref{table:results}, with further results in \cref{fig:app:tradeoff_grid_llama,fig:app:tradeoff_grid_misc,fig:app:tradeoff_grid_pythia}. 


We observe that \method{} performance is robust across all tasks and models tested, as compression ratios of $1/2$ to $1/8$ are readily achievable with little to no loss in task performance. H$_2$O can attain good performance on some tasks such as TriviaQA and WikiTest-$103$, although other tasks, including SQuAD and Text Repetition, are more challenging and notable degradation occurs. LM-Infinite performance degrades across all tasks, demonstrating that the tasks do not permit the trivial solution of discarding the long input sequence. 

\subsection{Sequence Length Scaling}
\begin{figure}[t]
\begin{center}
    \includegraphics[width=\linewidth]{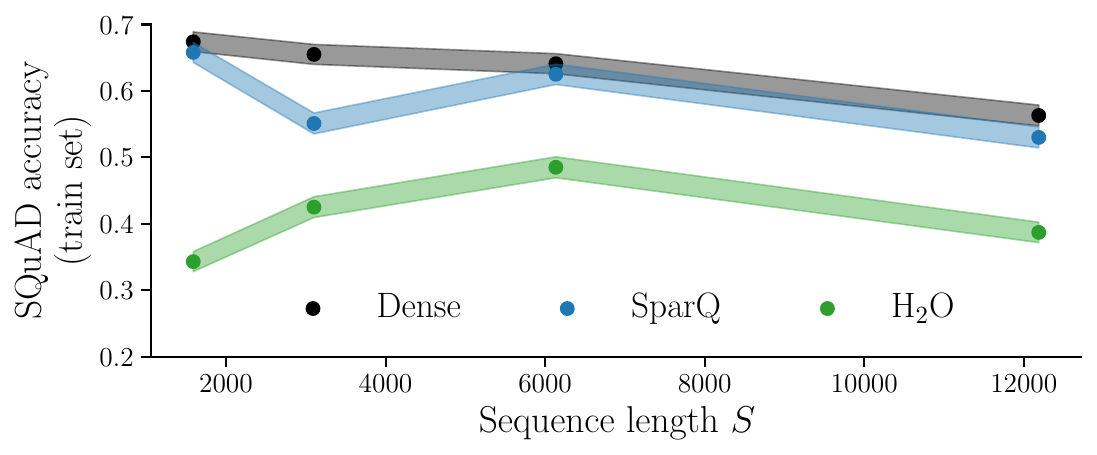}
    \caption{SQuAD performance vs input sequence length. The compression ratio is fixed at $1/4$. Uses Vicuna $1.5$ $7$B with $16$k maximum sequence length against our SQuAD (train) task with $7$ (default) to $63$ confusion contexts to increase the sequence length. We believe the drop in performance at $3$k tokens is an artefact of the RoPE scaling and fine-tuning procedure used to extend the context window of the Vicuna model.}
    \vspace*{-1em}
\label{fig:ablations:sequence_length_scaling}
\end{center}
\end{figure}
The sequence lengths of different examples in our main tasks vary between $1$k and $2$k tokens, whereas many LLMs support sequence lengths far greater than this. We considered two tasks to evaluate how \method{} performs as sequence length scales.

The first task is a variation of SQuAD, which increases both sequence length and task difficulty by increasing the number of confusion contexts present in the prompt, which is akin to increasing the number of retrieved documents with a retrieval augmented generation system \citep{borgeaud2022improving}. We test \method{} and H$_2$O in this setting using Vicuna \citep{vicuna2023}, a descendent of Llama $2$ that has been adapted for longer sequences. Both \method{} and H$_2$O are configured to maintain a fixed compression ratio versus the dense baseline (keeping $r=32$ and modifying $k$ to maintain $1/4$ compression). The results in \cref{fig:ablations:sequence_length_scaling} show that \method{} is scalable to large sequences, as it can maintain performance up to $128$k sequence length.

The second task evaluated is \emph{needle in a haystack}, in which a ``text needle'' is inserted into the context at a certain depth, and the model is tasked with retrieving information from the needle. The exact implementation of this task we used is outlined in \citet{dhinakaran2024needleinahaystack}. We compare \method{} with H$_2$O and LM-Infinite over a range of different compression ratios. The results, as seen in \cref{table:needle-in-a-haystack} and \cref{fig:app:niah_heatmaps}, show that \method{} achieves performance very close to the dense attention baseline, even in high-sparsity settings.

 

\begin{table}
\caption{Needle in a haystack results, averaged over all depths and sequence lengths $S$ within the specified range. See \cref{fig:app:niah_heatmaps} for full heatmap results and more details.}
  \label{table:needle-in-a-haystack}
\resizebox{\linewidth}{!}{%
{\renewcommand{\arraystretch}{1}
    \begin{tabular}{ ccccc }
\toprule
\textbf{$\bm{S}$ Range} & \textbf{Compression} & \textbf{LM-$\bm{\infty}$} & \textbf{H$_2$O} & \textbf{\methodshort{}} \\\midrule

\multirow{3}*{$8$k - $16$k} & $1$ & $\bm{100}$\textbf{\%} & $\bm{100}$\textbf{\%} & $\bm{100}$\textbf{\%}\\
& $1/4$ & $23.5$\% & $5.9$\% & $\bm{100}$\textbf{\%}\\
& $1/8$ & $11.8$\% & $2.9$\% & $\bm{79.4}$\textbf{\%}\\\midrule

\multirow{3}*{$16$k - $24$k} & $1$ & $\bm{100}$\textbf{\%} & $\bm{100}$\textbf{\%} & $\bm{100}$\textbf{\%}\\
& $1/4$ & $23.5$\% & $8.8$\% & $\bm{100}$\textbf{\%}\\
& $1/8$ & $11.8$\% & $5.9$\% & $\bm{100}$\textbf{\%}\\\midrule

\multirow{3}*{$24$k - $32$k} & $1$ & $\bm{90.4}$\textbf{\%} & $\bm{90.4}$\textbf{\%} & $\bm{90.4}$\textbf{\%}\\
& $1/4$ & $23.5$\% & $10.3$\% & $\bm{90.4}$\textbf{\%}\\
& $1/8$ & $11.8$\% & $5.9$\% & $\bm{87.5}$\textbf{\%}\\\bottomrule

\end{tabular}
}
}
\end{table}

\subsection{Ablations}

\begin{figure}[t]
    \begin{center}
    \subfloat[\label{fig:ablations:oracle_comparison}Accuracy results of \method{} and a random low rank compression scheme against an oracle top-$k$ selector.]{%
    \includegraphics[width=\columnwidth]{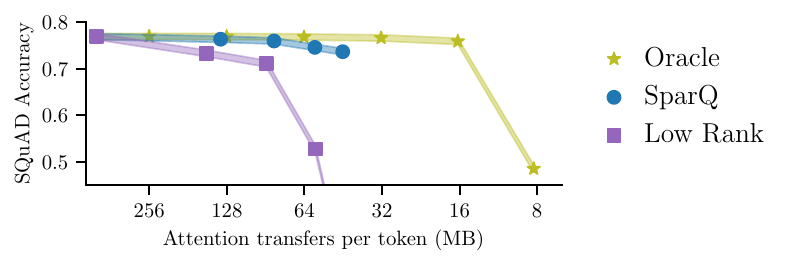}
    }
    \qquad
    \subfloat[\label{fig:ablations:temperatures}Comparison of different softmax temperatures for approximate attention scores for two different hyperparameter configurations.]{%
    \includegraphics[width=\columnwidth]{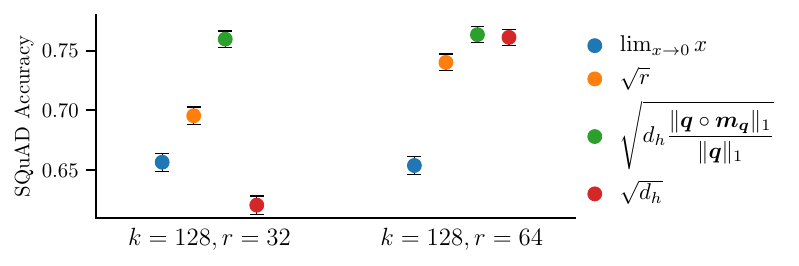}
    }
    \qquad
    \subfloat[\label{fig:ablations:recipe}Results for Repetition and SQuAD tasks with $r \in \{16,32,64\}$.]{%
    \includegraphics[width=\columnwidth]{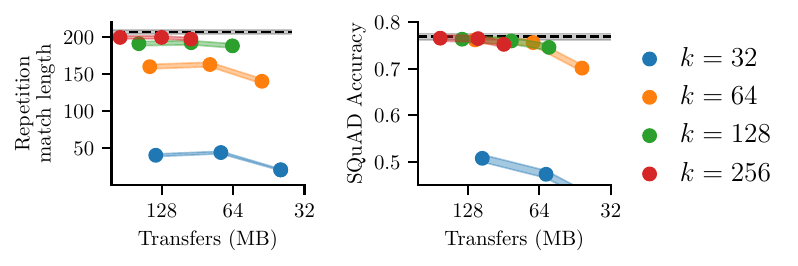}
    }
    \end{center}
    \caption{Ablations results, investigated on Llama $2$ $7$B.}
\end{figure}

\paragraph{Key cache compression} The first step in \method{} involves reading $r$ components of the key cache to approximately determine which keys yield the highest attention scores. To examine the practical trade-off of the approximation, we look at how \method{} performs when compared to a theoretical upper-bounding ``\emph{oracle}'' which provides the exact top-$k$ keys without requiring any data transfer to calculate the top-$k$. The results in \cref{fig:ablations:oracle_comparison} show that \method{} retains comparable performance to the oracle for a wide range of compression ratios, and attains considerably higher performance than a baseline compression scheme, in which a random low rank projection of $\bm{K}$ is transferred from memory.

\paragraph{Approximate softmax temperature} To empirically support our statistical analysis of $\alpha$ agreement shown in \cref{fig:approximation_analysis:reallocation_scale_scatter}, we evaluate a number of different viable temperature settings, including the square root of the head dimension ($\tau=\sqrt{d_h}$), the square root of the rank ($\tau=\sqrt{r}$), and the temperature proposed in \cref{eq:l1coveragetemp}. We also consider the scenario where we do not reallocate mass to mean value ($\alpha=0$), which corresponds to the limit of the temperature tending towards $0$. We find that our proposed temperature performs best, as shown in \cref{fig:ablations:temperatures}.

\paragraph{Hyperparameter selection} The reduction of data transfer attained by \method{} is controlled by its two hyperparameters, $k$ and $r$. Reducing either of these variables will improve the bandwidth efficiency, but can negatively impact task performance. \cref{fig:ablations:recipe} shows the relationship between $k$ and $r$ on both of these factors. Based on these results, we propose a simple recipe of setting $k=128$ and tuning $r$ to maintain a good trade-off between data transfer and task performance for a range of models and tasks.

\section{Benchmarking}
\label{sec:benchmarking}

The results above use a theoretical cost model of total memory transfers (the number of scalar elements transferred to and from memory per token), allowing us to evaluate SparQ Attention independently of the specific hardware setup and number formats used. To validate our findings, we performed a set of microbenchmarks of an attention operation in isolation, in addition to end-to-end performance benchmarks.

SparQ Attention benefits from two optimisations. The first is to store $\bm{K}$ twice, in both $d_h$-contiguous and $S$-contiguous layouts, since this allows for an efficient gather (indexing) on either axis, at the cost of $50\%$ extra memory usage. The second optimisation is to use a fused gather-then-matmul operation to avoid writing the result of the gather to memory.

\subsection{Microbenchmarks}
We tested multiple implementations of baseline and SparQ Attention on IPU using the Poplar C++ interface and GPU using PyTorch \citep{paszke2019pytorch}. In all cases, we used the Llama $2$ $7$B shape parameters: $32$ heads, $d_h=128$. The implementations tested were: \textbf{Dense} baseline, choosing the faster of a plain PyTorch implementation and the builtin \texttt{scaled\_dot\_product\_attention}, \textbf{SparQ (Triton)}, storing $\bm{K}$ twice and using fused gather-then-matmul kernels written using Triton \citep{tillet2019triton}, \textbf{SparQ (PyTorch)}, with no Triton and \textbf{SparQ (Triton, $1\!\times\!\boldsymbol{K}$)}, storing $\bm{K}$ in $d_h$-contiguous layout only, for no additional memory cost.
\begin{figure}[t]
\begin{center}
    \includegraphics[width=\linewidth]{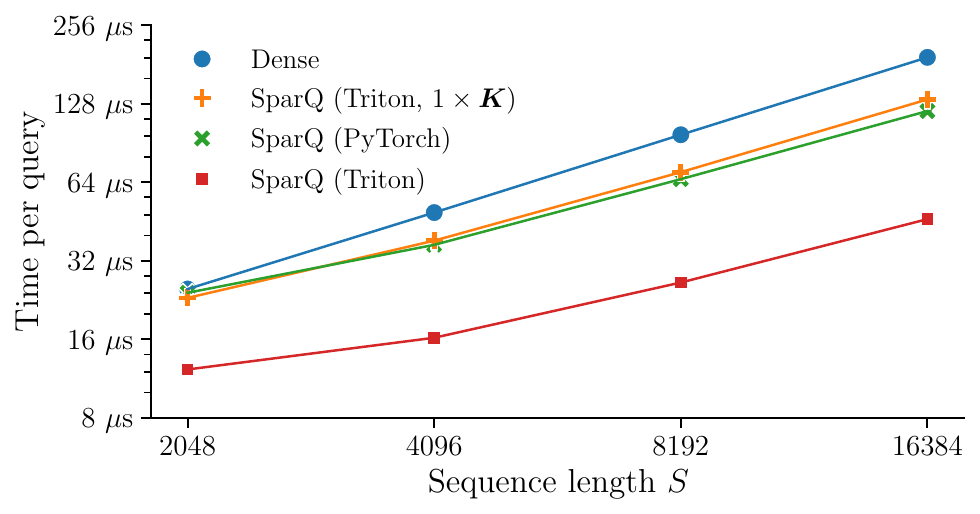}
    \caption{Microbenchmark results for batch size $64$, $32$ heads, $d_h=128$, $r=32$, $k=128$, A$100$ ($40$GB).}
\label{fig:benchmark_a100}
\end{center}
\end{figure}
In an example configuration running on a single IPU from a Bow Pod$_{16}$, batch size $1$, sequence length $S=16384$, the dense baseline achieves 40.4 ms/query, while SparQ $(r=32$, $k=128$) achieves $5.28$ ms/query for a speedup of $7.41\times$ (the theoretical speedup of SparQ is $7.53\times$). This near-perfect speedup is achieved because attention is strongly memory bound when using remote memory. In contrast, the baseline running in local SRAM takes $134$ \us{} for a $345\times$ speedup, but this is only practically achievable when the whole model fits in SRAM.

\begin{table}[t]
\caption{GPU microbenchmark performance with batch size $64$, sequence length $S=4096$, $r=32$, $k=128$. The theoretical speed-up is $6.4\times$.}
\label{tab:gpu_speedup}
\centering
\resizebox{\columnwidth}{!}{%
\begin{tabular}{ ccc }
\toprule
\textbf{Kernel} & \textbf{A100 (40GB)} & \textbf{A10G} \\\midrule

Dense
    & 49 \us{} ($1\times$)
    & 128 \us{} ($1\times$)
    \\

SparQ (Triton, $1\!\times\!\bm{K}$)
    & 38 \us{} ($1.28\times$)
    & 79 \us{} ($1.63\times$)
    \\

SparQ (PyTorch)
    & 37 \us{} ($1.33\times$)
    & 78 \us{} ($1.63\times$)
    \\

SparQ (Triton)
    & $\mathbf{16}$ \boldus{} ($\mathbf{3.02\times}$)
    & $\mathbf{31}$ \boldus{} ($\mathbf{4.17\times}$)
    \\\bottomrule

\end{tabular}

}
\end{table}

Our achieved GPU speed-ups are presented in \cref{tab:gpu_speedup}, and the performance trend with sequence length is shown in \cref{fig:benchmark_a100}. Standard error for all results given is $<1\%$ of the mean. See \cref{sec:app:benchmarking} for further details.

These microbenchmark results show that the theoretical benefits of SparQ Attention can yield substantial wall-clock time speedups on current hardware. Further work is needed to show improvements for small batch sizes, and to investigate alternatives to storing $\bm{K}$ twice.

\subsection{End-to-End Performance}

\begin{figure}[t]
\begin{center}
    \includegraphics[width=\linewidth]{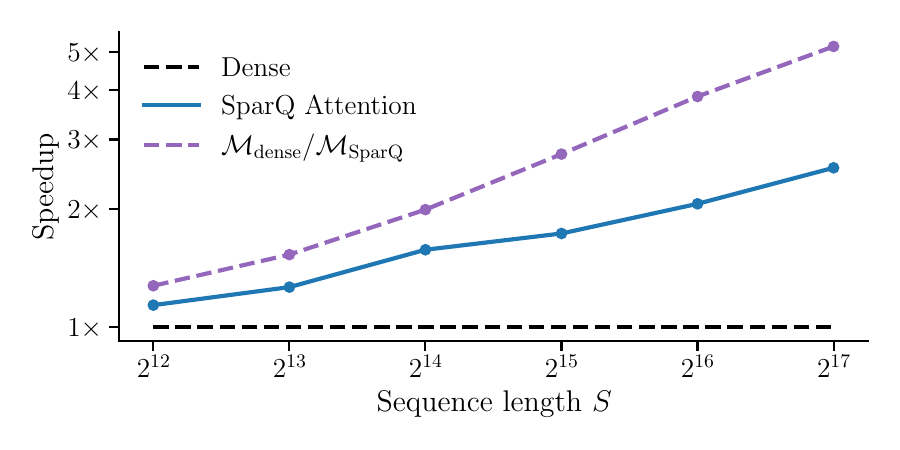}
    \vspace*{-1.5em}
    \caption{End-to-end CPU speedup results for \method{} compared to the dense baseline with Llama $2$ $7$B. This was achieved with batch size $1$ and a compression ratio of $1/8$, with model weights represented in $8$-bits and the KV cache represented in $16$-bits. The $\mathcal{M}_\mathrm{dense} / \mathcal{M}_\mathrm{SparQ}$ line is a theoretical result, illustrating the upper bound of speedups attained with \method{}.}
\label{fig:sparq_speedup}
\end{center}
\end{figure}

In addition to the positive microbenchmark results, we further highlight the practical improvements \method{} offers by benchmarking performance of the entire Transformer model on both CPU and GPU, implemented in \texttt{llama.cpp} and \texttt{gpt-fast} \citep{pytorch2023gptfast} respectively. In both cases, we measured the time it took to generate a single token, given an existing sequence length $S$.

\paragraph{CPU benchmarking}We evaluated CPU benchmarking performance on AMD EPYC systems with up to $256$GB memory. The results, as seen in \cref{fig:sparq_speedup}, show that \method{} attains speedups at all sequence lengths evaluated, compared to the dense baseline. At the longest sequence lengths considered, \method{} achieves a $2.5\times$ speedup, showcasing the benefits of reducing the data transfer associated with attention.

\begin{figure}[t]
\begin{center}
    \includegraphics[width=\linewidth]{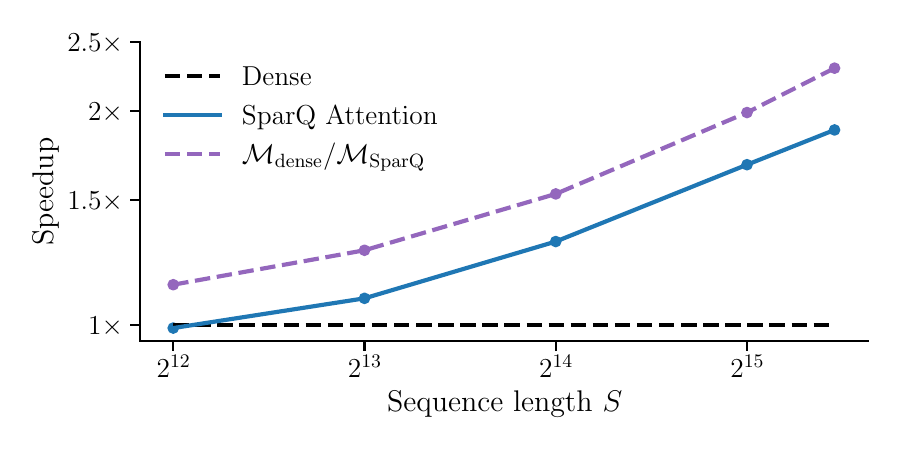}
    \vspace*{-1.5em}
    \caption{End-to-end H100 speedup results for \method{} compared to the dense baseline with Llama $2$ $7$B. These speedups were achieved with batch size $1$ and a compression ratio of $1/8$, with both model weights and KV cache represented in $16$ bits. The $\mathcal{M}_\mathrm{dense} / \mathcal{M}_\mathrm{SparQ}$ illustrates the theoretical maximum performance attainable under this setting.}
\label{fig:sparq_h100_speedup}
\end{center}
\end{figure}
\paragraph{GPU benchmarking} Our end-to-end GPU implementation is evaluated on a single H$100$ PCIe with $80$GB memory. Despite utilising high-bandwidth memory, GPU inference still achieves end-to-end speedups when using \method{} on modest sequence lengths, as seen in \cref{fig:sparq_h100_speedup}.

\section{Related Work}
\label{sec:related_work}
Efficient attention methods have been a very active area of research \citep{attention_review}. Schemes such as \emph{Sparse Transformers} \citep{child2019sparsetransformers}, \emph{Combiner} \citep{ren2021combiner}, \emph{Longformer} \citep{beltagy2020longformer}, \emph{BigBird} \citep{zaheer2020big}, \emph{Reformer} \citep{kitaev2020reformer} and \emph{Sparse Sinkhorn Attention} \citep{tay2020sinkhorn} have been developed to increase efficiency of the attention mechanism by extracting information from the most salient tokens in the sequence or approximating dense attention maps. Two schemes that reduce memory footprint and data transfer of the attention operation, while maintaining quadratic complexity are \emph{Multi-Query Attention} (MQA) \citep{shazeer2019fast} and \emph{Grouped-Query Attention} (GQA) \citep{ainslie2023gqa} that share each KV head across multiple query heads. These methods form part of the architecture: they must be implemented during pre-training, carry varying task performance trade-offs, and may affect model quality and stability.

An emerging area of research similar to \method{} aims to only adapt the inference procedure of a pre-trained model. The simplest method of this category is part of \emph{FlexGen} \citep{sheng2023flexgen}, and calculates exact attention scores, retrieving only the values associated with the top-$k$ scores. This process uses the full key cache to produce attention scores, limiting the asymptotic reduction of the memory transfers to only $50\%$. LM-Infinite \citep{han2023lminfinite} and StreamingLLM \citep{xiao2023streamingllm} employ a fixed sparsity pattern preserving the most recent tokens and a few initial tokens for better attention efficiency, but are not selective in their cache lookup.

\emph{Eviction} schemes cache only a subset of keys and values, by continually deleting tokens that are uninformative for future outputs. By reducing the cache size itself, both the amount of memory used and data transferred are reduced. \emph{H$_2$O} \citep{zhang2023h2o}, \emph{Scissorhands} \citep{liu2023scissorhands} and \emph{FastGen} \citep{ge2024fastgen} are examples of such eviction methods. H$_2$O uses a greedy eviction policy that maintains in memory the most salient ``Heavy Hitter'' tokens  that  contribute most to the attention scores. Scissorhands identifies and maintains ``pivotal tokens'' by counting when a token's attention score exceeds an importance threshold. FastGen adopts heuristics such as preventing the eviction of special tokens and punctuation, and tailors the compression strategy to each individual attention head. While these methods reduce the memory footprint of the KV cache as well as data transfer, they also lead to permanent loss of information from the context window, which can lead to mistakes for queries seeking less-attended parts of the sequence.

\emph{IceFormer} \citep{iceformer} uses multiple existing approximate nearest neighbour algorithms for approximating attention scores of pre-trained models, focusing on speeding-up the prefill stage, rather than generation. \emph{Scatterbrain} \citep{chen2021scatterbrain} employs similar techniques, but for computer vision applications.

In addition to compressing the KV cache, a number of methods strive to speed up LLM inference by inducing sparsity in the weights of the model. Deja Vu \citep{liu2023dejavu} is a contextually-sparse approach that aims to predict which model parameters are required such that the error between the full computation and sparse approximation is minimised. Similarly, activation sparsity methods, including \citet{kurtz2020activation_sparsity} and \citet{mirzadeh2024relufication}, exploit zero-values found in activations, typically induced by ReLU activation functions. \citet{kurtz2020activation_sparsity} introduce an alternative \emph{forced activation threshold ReLU} function which can induce sparsity at specified thresholds. Similarly, \citet{mirzadeh2024relufication} replace the activation functions in LLMs with ReLUs, followed by additional fine-tuning. These methods are most suitable for small batch size and short sequence length regimes, where inference is bottlenecked by parameter transfer, rather than the KV cache, but are compatible with sparse attention techniques such as \method{}.

An orthogonal line of work increases bandwidth efficiency by compressing the KV cache with $4$-bit number formats \citep{liu2023llmqat, sheng2023flexgen}. \citet{liu2023scissorhands} demonstrate that $4$-bit compression is complementary to techniques that reduce the number of transferred elements.




\section{Conclusion}
In this work we have presented \method{}, a novel technique for unlocking faster inference for pre-trained LLMs. Our proposed technique modifies the attention mechanism to access only the relevant tokens from the KV cache at every generation step, leading to considerable data transfer savings. This is particularly beneficial in long sequence length regimes, where inference speed is often bottlenecked by memory transfers rather than computation.

We also highlight the advantages of maintaining the full KV cache in memory for task performance by comparing \method{} to other popular strategies which discard information from the input sequence. These alternative approaches rely on heuristics or predefined policies to determine which items in the KV cache to remove, which may not generalise across the wide range of applications LLMs are used for. We show that \method{} is robust across numerous tasks and models, making it a viable technique for reducing inference times in unseen settings. 

%


\section*{Impact Statement}

This paper presents work whose goal is to advance the field of Machine Learning. There are many potential societal consequences of our work, none which we feel must be specifically highlighted here.


\section*{Acknowledgements}

We would like to thank Oscar Key for implementing \method{} on GPU and benchmarking its end-to-end performance. 

In addition, we would also like to thank Daniel Justus, Paul Balan\c{c}a and Andrew Fitzgibbon for their helpful input and feedback on this work.

\bibliography{main}
\bibliographystyle{icml2024}


\newpage
\appendix
\onecolumn

\section{Detailed Results}
\label{sec:app:results}

\renewcommand\thefigure{\ref{sec:app:results}\arabic{figure}}
\setcounter{figure}{0}
\renewcommand\thetable{\ref{sec:app:results}\arabic{table}}
\setcounter{table}{0}
\renewcommand\theequation{\ref{sec:app:results}\arabic{equation}}
\setcounter{equation}{0}

\begin{figure}[H]
    \centering
    \includegraphics[height=19cm]{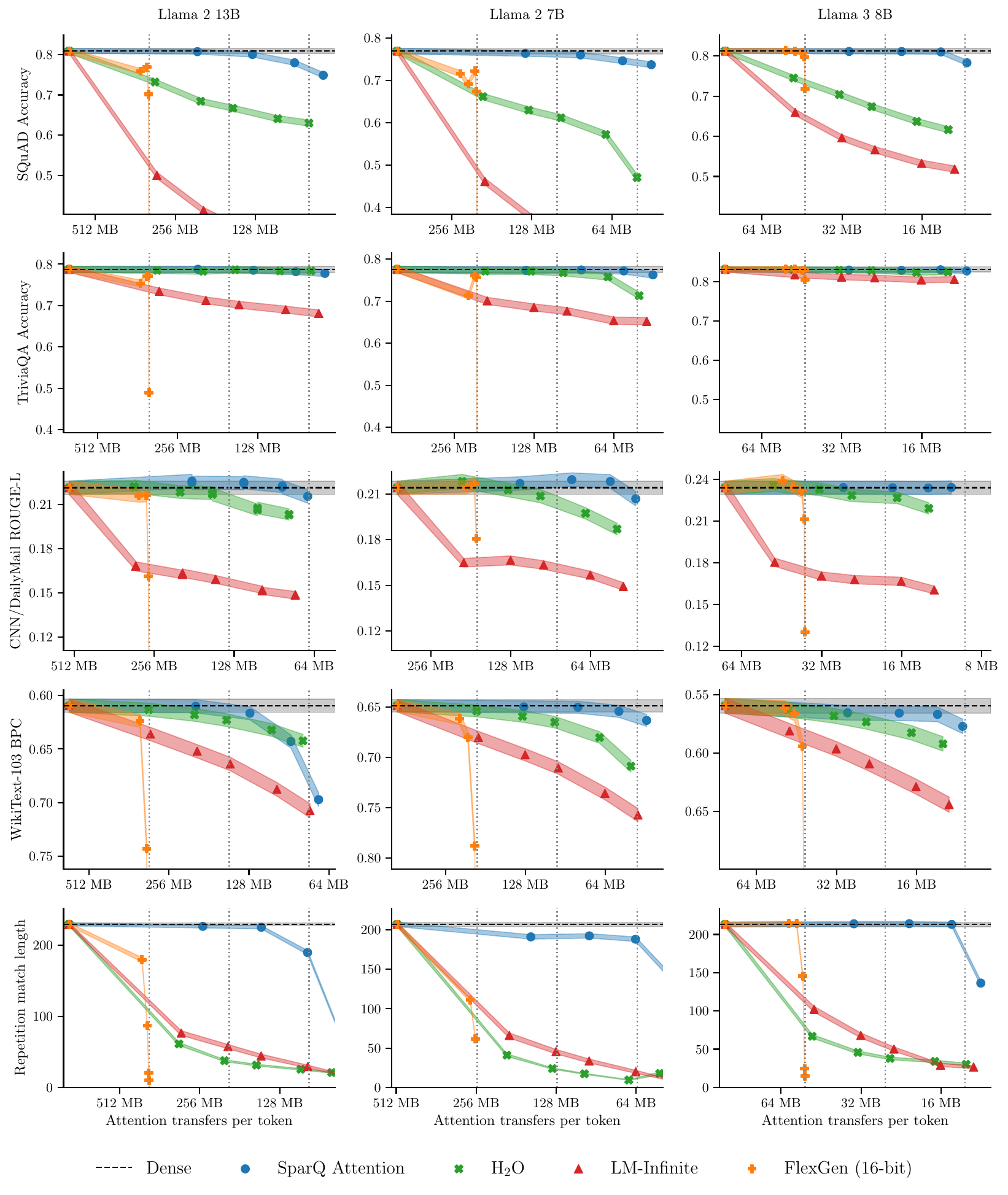}
    \caption{Compression versus performance trade-off curves over all tasks for the Llama $2$ and Llama $3$ model families. The y-axis minimum is set to ($0.5$, $0.5$, $0.5$, $1.25$, $0.0$)$\times$ the dense baseline for the tasks, reading top-to-bottom, in order to give a consistent view of the performance loss across models. Vertical dotted lines show $1/2 \times$, $1/4 \times$ and $1/8 \times$ compression versus dense. Shaded lines show $\pm 1$ standard error of the mean (uncertainty due to a finite test set).}
    \label{fig:app:tradeoff_grid_llama}
\end{figure}

\begin{figure}[H]
    \centering
    \includegraphics[height=19cm]{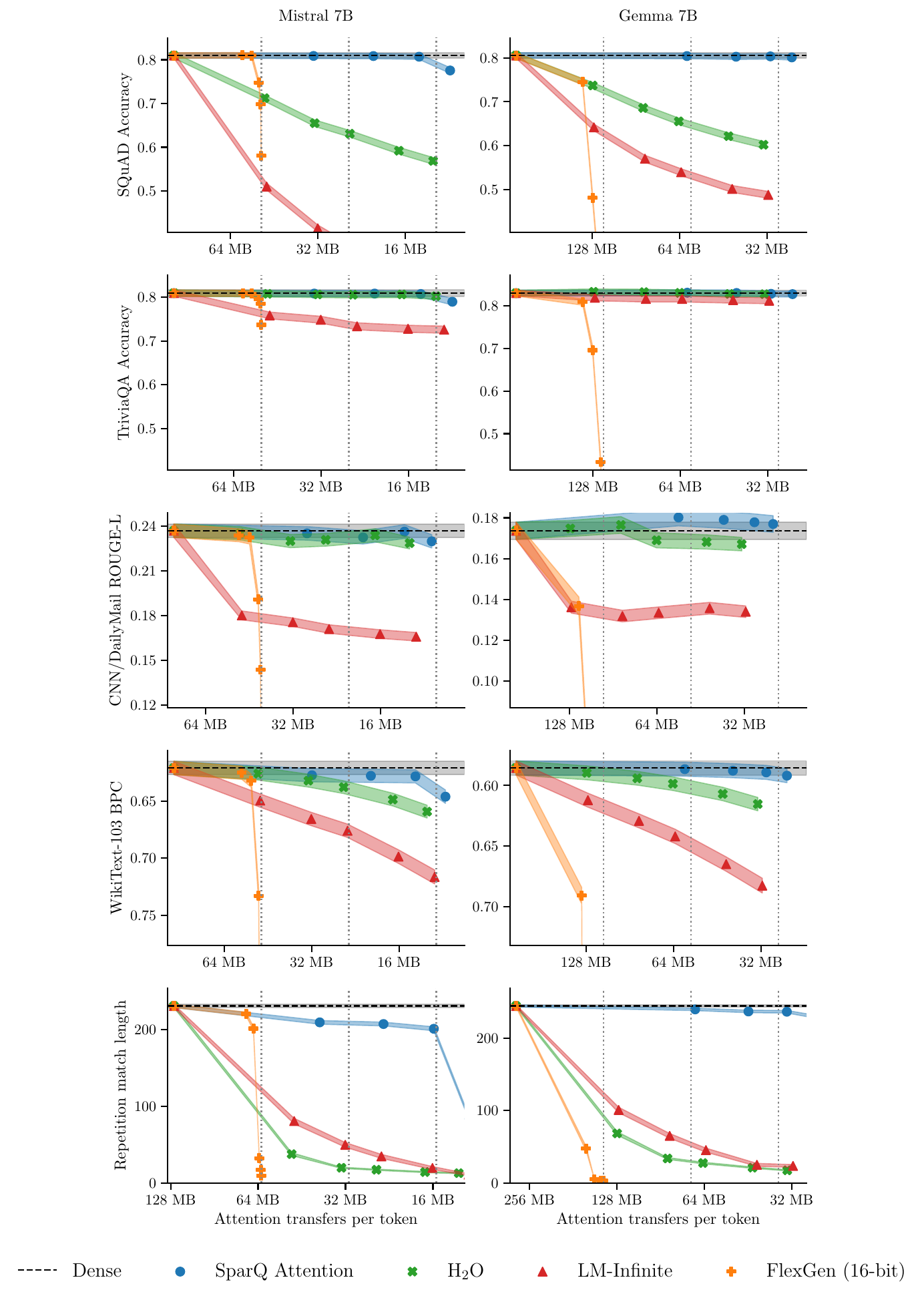}
    \caption{Compression versus performance trade-off curves over all tasks for Mistral $7$B and Gemma $7$B models. The y-axis minimum is set to ($0.5$, $0.5$, $0.5$, $1.25$, $0.0$)$\times$ the dense baseline for the tasks, reading top-to-bottom, in order to give a consistent view of the performance loss across models. Vertical dotted lines show $1/2 \times$, $1/4 \times$ and $1/8 \times$ compression versus dense. Shaded lines show $\pm 1$ standard error of the mean (uncertainty due to a finite test set).}
    \label{fig:app:tradeoff_grid_misc}
\end{figure}

\begin{figure}[H]
    \centering
    \includegraphics[height=19cm]{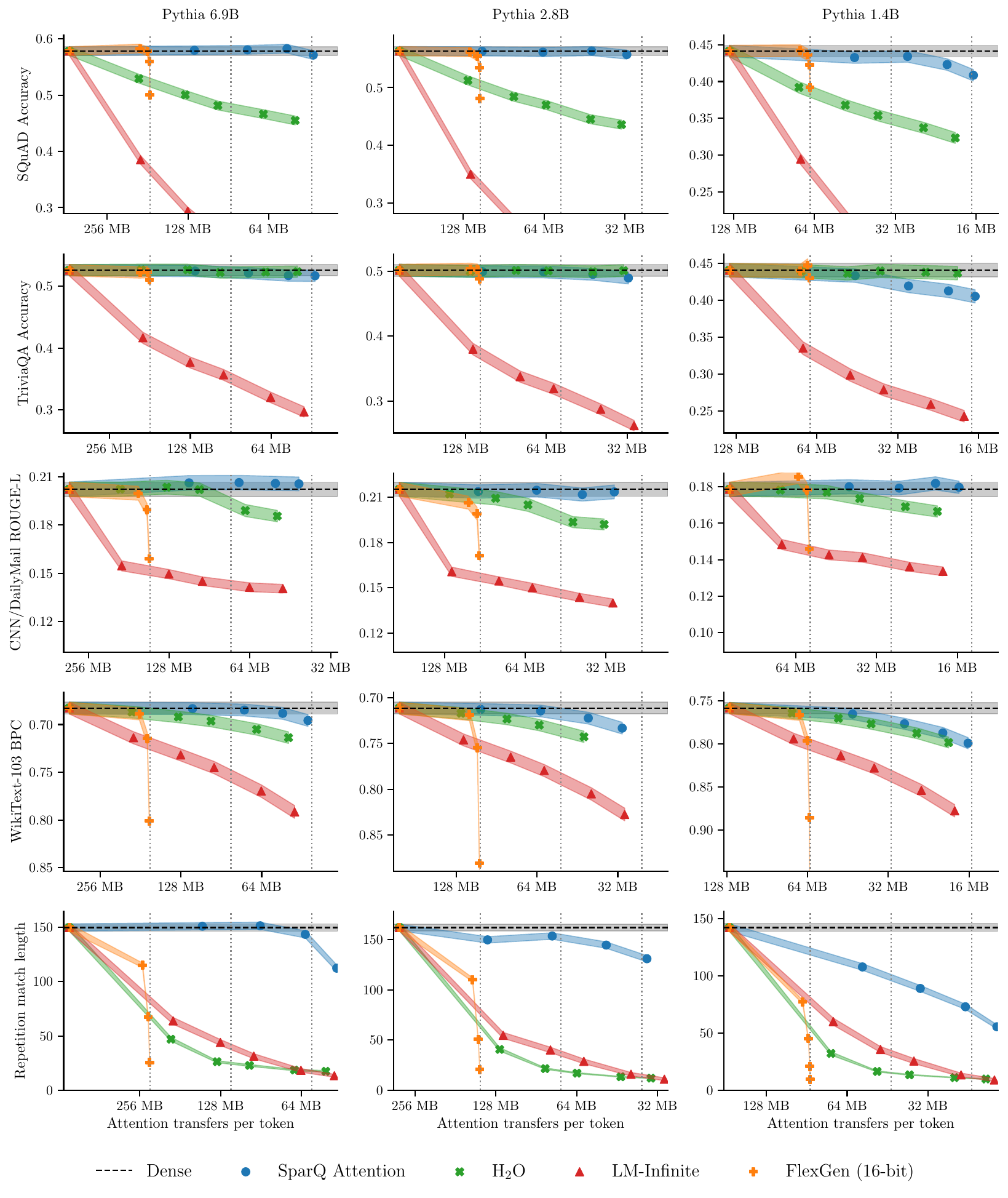}
    \caption{Compression versus performance trade-off curves over all tasks for the Pythia family of models. The y-axis minimum is set to ($0.5$, $0.5$, $0.5$, $1.25$, $0.0$)$\times$ the dense baseline for the tasks, reading top-to-bottom, in order to give a consistent view of the performance loss across models. Vertical dotted lines show $1/2 \times$, $1/4 \times$ and $1/8 \times$ compression versus dense. Shaded lines show $\pm 1$ standard error of the mean (uncertainty due to a finite test set).}
    \label{fig:app:tradeoff_grid_pythia}
\end{figure}

\Cref{fig:app:tradeoff_grid_llama,fig:app:tradeoff_grid_misc,fig:app:tradeoff_grid_pythia} report the compression/performance trade-off curves for all models and tasks that were evaluated.

\vspace*{\fill}
\newpage
\vspace*{\fill}
\vspace*{\fill}

\newpage
\vspace*{\fill}
\vspace*{\fill}

\newpage

\begin{figure}[H]
        \centering
        \includegraphics[width=7cm]{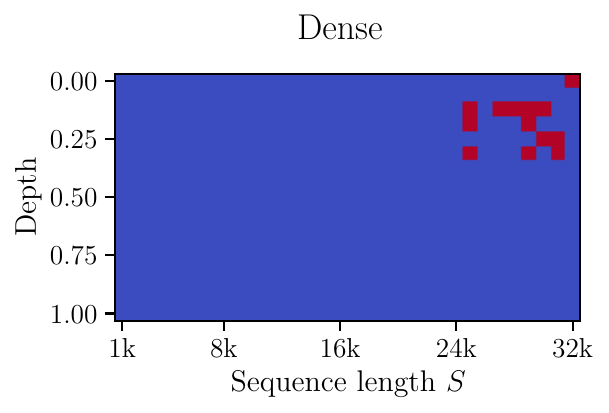}\vspace*{1em}
        \includegraphics[width=13cm]{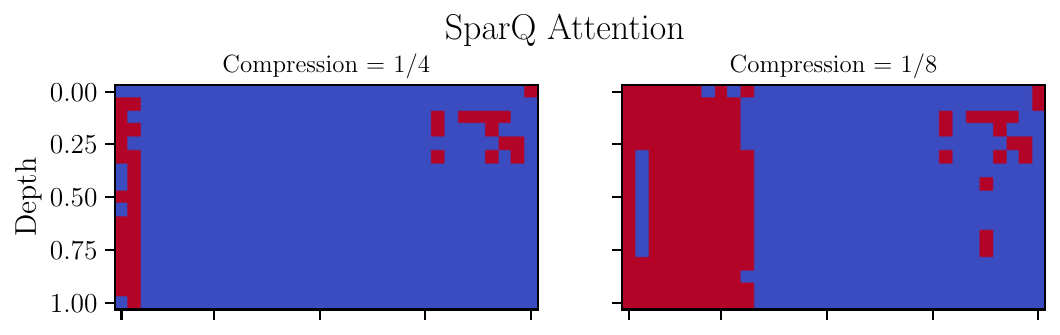}\vspace*{1em}
        \includegraphics[width=13cm]{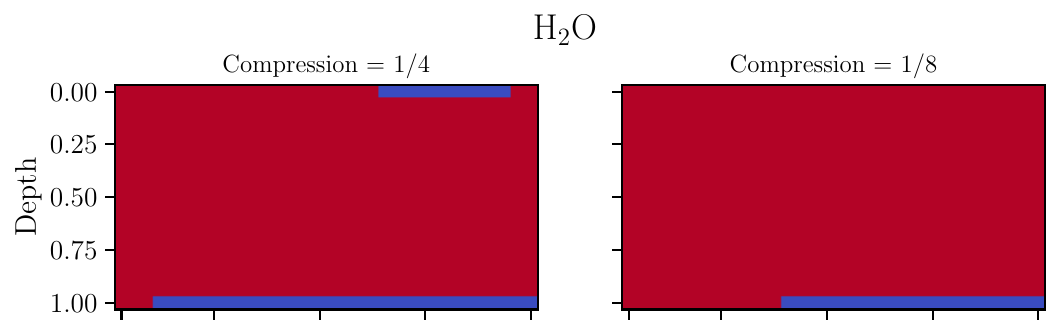}\vspace*{1em}
        \includegraphics[width=13cm]{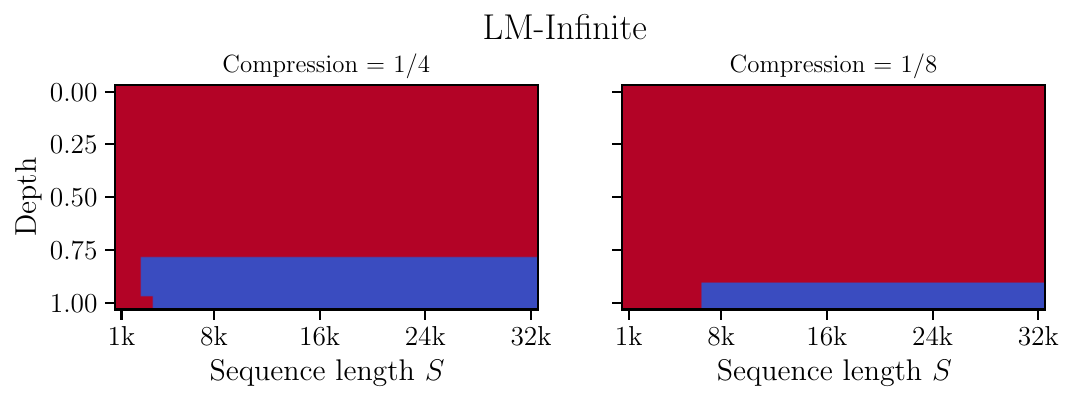}
    \caption{Heatmaps of the results for the needle in a haystack task, as outlined in \citet{dhinakaran2024needleinahaystack}. For various sequence lengths (comprising of essays by Paul Graham), the needle (which is the sequence ``\textit{The best thing to do in San Francisco is eat a sandwich and sit in Dolores Park on a sunny day}'') is inserted into the sequence at various depths, and the model is prompted to answer what the best thing to do in San Francisco is. The heatmaps show whether the model returns the correct (blue) or incorrect (red) answer. We evaluated this task over \method{}, H$_2$O and LM-Infinite (in addition to the dense baseline) on the \texttt{togethercomputer/LLaMA-2-7B-32K} model, for two compression ratios.}
    \label{fig:app:niah_heatmaps}
\end{figure}

\section{Code}
\label{sec:app:code}
\begin{minted}{python3}
from torch import softmax, sqrt, tensor, topk


def gather(t, dim, i):
    dim += (dim < 0) * t.ndim
    return t.gather(dim, i.expand(*t.shape[:dim], i.shape[dim], *t.shape[dim + 1 :]))


def attn(Q, K, V, M):
    s = (Q @ K.transpose(-1, -2)) / sqrt(tensor(Q.shape[-1])) + M
    y = softmax(s, dim=-1) @ V
    return y


def sparq_attn(Q, K, V, V_mean, M, r, k):
    # Q -- (batch_size, n_kv_heads, n_heads // n_kv_heads, 1, head_size)
    # K, V -- (batch_size, n_kv_heads, 1, seq_len, head_size)

    # 1. Approximate attention scores using r largest components of Q
    i1 = topk(abs(Q).sum(dim=2, keepdim=True), r, -1).indices
    Q_hat, K_hat = gather(Q, -1, i1), gather(K, -1, i1)
    scale = sqrt(
        Q.shape[-1]
        * abs(Q_hat).sum(dim=-1, keepdim=True)
        / abs(Q).sum(dim=-1, keepdim=True)
    )
    s_hat = softmax(Q_hat @ K_hat.transpose(-1, -2) / scale + M, dim=-1)

    # 2. Gather top k positions based on approximate attention scores & run attention
    i2 = topk(s_hat.sum(dim=2, keepdim=True), k, -1).indices
    iKV = i2[..., 0, :, None]
    K, V, M = gather(K, -2, iKV), gather(V, -2, iKV), gather(M, -1, i2)
    y_ = attn(Q, K, V, M)

    # 3. Estimate the total score of the top k, and interpolate with V_mean
    alpha = gather(s_hat, -1, i2).sum(-1, keepdim=True)
    y = alpha * y_ + (1 - alpha) * V_mean
    return y
\end{minted}

\section{Arithmetic Intensity}
\label{sec:app:arithmetic_intensity}

\renewcommand\thefigure{\ref{sec:app:arithmetic_intensity}\arabic{figure}}
\setcounter{figure}{0}
\renewcommand\thetable{\ref{sec:app:arithmetic_intensity}\arabic{table}}
\setcounter{table}{0}
\renewcommand\theequation{\ref{sec:app:arithmetic_intensity}\arabic{equation}}
\setcounter{equation}{0}

Consider a full transformer layer, with $N$ parameters, batch size $B$, $C$ elements in the attention KV cache per batch element and $g$ grouped-query heads per key-value head. This implies the arithmetic intensity:
\begin{equation}
    \mathcal{\frac{A}{M}} = \frac{BN + BCg}{N + BC} = \frac{N + C g}{N/B + C} \label{eq:arithmetic_intensity_1}
\end{equation}
We can increase arithmetic intensity by making $B$ large, causing $\mathcal{A}/\mathcal{M}$ to approach $N/C + g$. Hence the limiting factor for large-batch transformer inference is the ratio of the KV cache size per-item to the size of the model.

We can alternatively express this in terms of the model's basic hyperparameters. A standard transformer with model dimension $d_m$ and sequence-length $S$ has $N = 12 (d_m)^2$ and $C = 2 \, S \, d_m / g$ \citep{kaplan}. Substituting these values into \cref{eq:arithmetic_intensity_1}, we get
\begin{equation}
    \mathcal{\frac{A}{M}} = \frac{6 + \rho \, g}{6 / B + \rho}
    \label{eqn:arithmetic_intensity_rho}
\end{equation}
where $\rho = S / (g d_m)$ is a variable we have introduced, and underlies the KV cache-model size relationship outlined above, determining the point at which the model becomes memory bandwidth bound. We observe that the arithmetic intensity as batch size increases approaches $g + 6/\rho$.

\begin{figure}[t]
    \centering
    \subfloat[\label{fig:app:arithmetic_intensity:g1}]{
    \includegraphics[width=8cm]{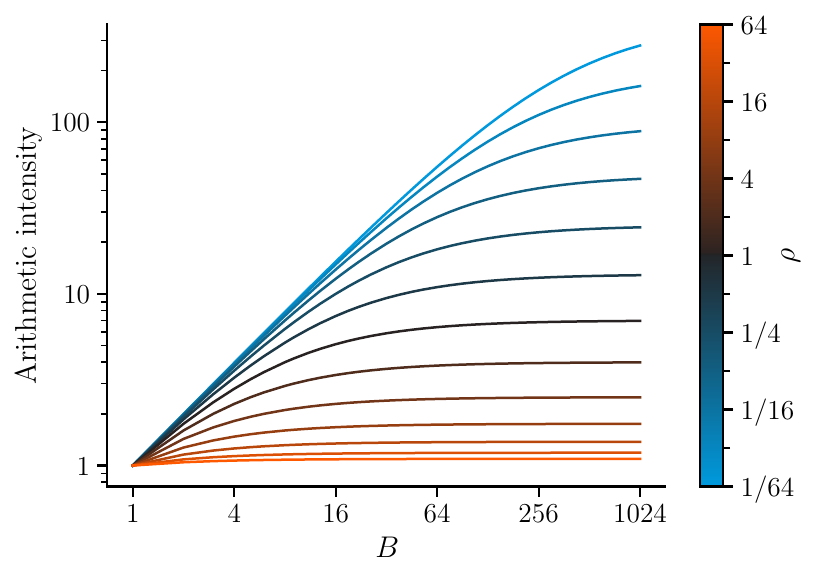}
    }
    \subfloat[\label{fig:app:arithmetic_intensity:g8}]{
    \includegraphics[width=8cm]{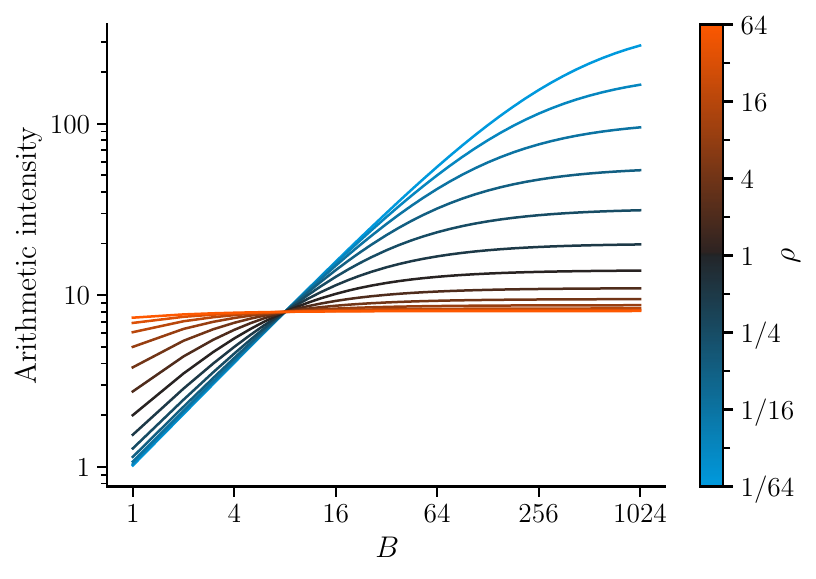}
    }
    \caption{Relationship between $\rho\!=\!S/(g d_m)$, $B$ and arithmetic intensity. \subref{fig:app:arithmetic_intensity:g1} Multi-head attention, $g=1$. \subref{fig:app:arithmetic_intensity:g8} Grouped-query attention, $g=8$. This highlights the importance of $\rho$, even with large batch size and GQA.
    }
    \label{fig:app:arithmetic_intensity}
\end{figure}

\begin{table}[H]
\centering
\begin{tabular}{cccccc}
    \toprule
    Model & $g$ & $d_m$ & $S$ & $\rho=S/(g d_m)$ & Max $\mathcal{A}/\mathcal{M}$ \\\midrule
    Llama~2 $7$B & $1$ & $4096$ & $4096$ & $1$ & $7$ \\
    Llama~2 $70$B & $8$ & $8192$ & $4096$ & $1/16$ & $104$ \\
    Llama~2 $70$B & $8$ & $8192$ & $16384$ & $1/4$ & $32$ \\
    \bottomrule
\end{tabular}
\end{table}
The general relationship between $\rho$ and arithmetic intensity is shown in \cref{fig:app:arithmetic_intensity}.

\paragraph{Hardware}

Properties of selected machine learning hardware.\footnote{For IPU \citep{bowIPU}, we use the exchange memory bandwidth of $11$~TB/s. A10 \citep{nvidiaA10}. H100 \citep{nvidiaH100}.} Note that $r_{\mathcal{A}}$ is the number of multiply-adds per second and $r_{\mathcal{M}}$ the number of data elements transferred per second.
\begin{table}[H]
\centering
\begin{tabular}{ccccc}
    \toprule
    Name & Memory technology & $r_{\mathcal{A}} /10^{12}$ & $r_{\mathcal{M}} /10^{12}$ & $r_{\mathcal{A}}/r_{\mathcal{M}}$ \\\midrule
    Bow IPU (FP$16$)
    & SRAM & $175$ & $5.5$ & $32$ \\
    A10 GPU (INT$8$)
    & GDDR & $125$ & $0.6$ & $210$ \\
    H100 SXM GPU (FP$8$)
    & HBM & $990$ & $3.35$ & $295$ \\
    \bottomrule
\end{tabular}
\end{table}
Comparing $r_{\mathcal{A}}/r_{\mathcal{M}}$ for this hardware to the arithmetic intensity achievable for standard transformer models, it's clear that sequence generation will hit a data transfer bottleneck.

In summary, we have seen that sequence generation exhibits a large-batch arithmetic intensity of just $7$ for multi-head attention with $S = d_m$, up to $100$ for grouped-query attention with $S \ll d_m$, while ML hardware can provide $r_{\mathcal{A}}/r_{\mathcal{M}} > 200$.

\section{Measuring Time Spent in Attention}
\label{sec:app:llama.cpp}

\renewcommand\thefigure{\ref{sec:app:llama.cpp}\arabic{figure}}
\setcounter{figure}{0}
\renewcommand\thetable{\ref{sec:app:llama.cpp}\arabic{table}}
\setcounter{table}{0}
\renewcommand\theequation{\ref{sec:app:llama.cpp}\arabic{equation}}
\setcounter{equation}{0}

\begin{figure}[h]
    \centering
    \subfloat[\label{fig:app:time_to_generate_tokens_llama_cpp:cpu}]{
    \includegraphics[width=8cm]{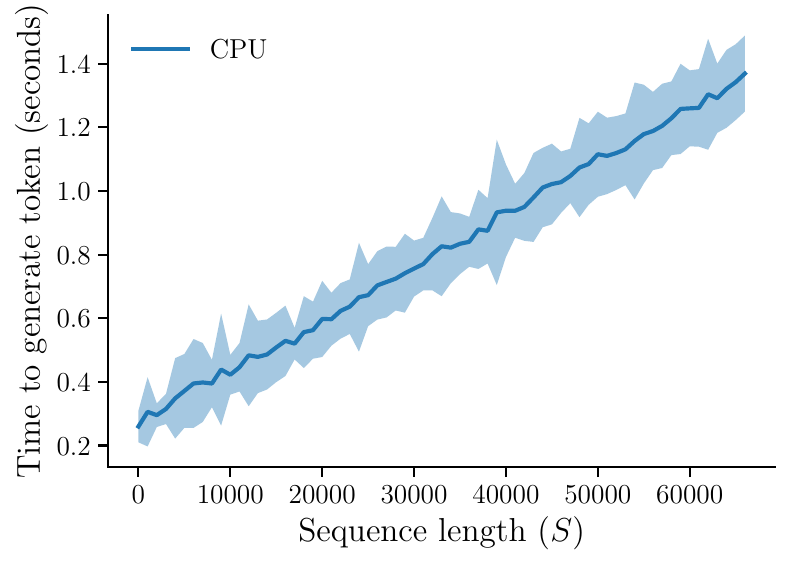}
    }
    \subfloat[\label{fig:app:time_to_generate_tokens_llama_cpp:gpu}]{
    \includegraphics[width=8cm]{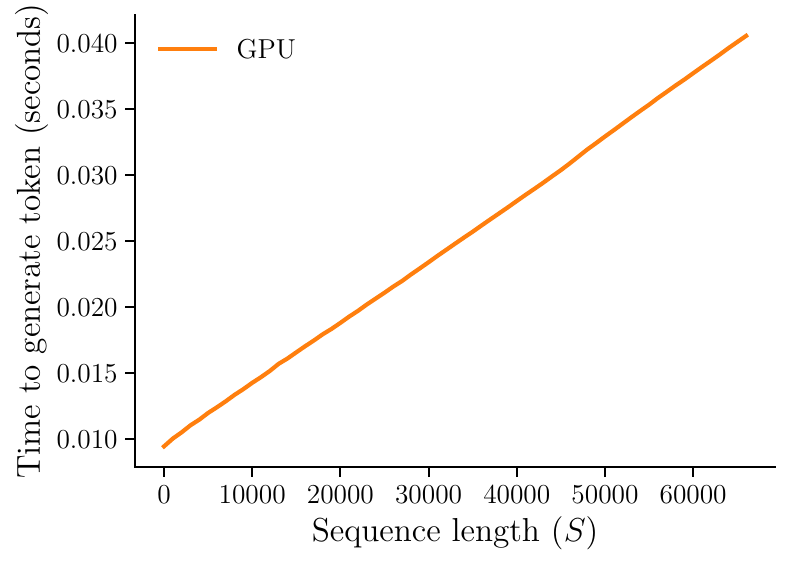}
    }
    \caption{For each sequence length $S$, we prefill the KV cache with a prompt of length $S$ before measuring the time it takes to generate a single additional token with Llama $2$ $7$B (batch size $1$). We report the mean and standard deviation over $200$ runs per sequence length, for \subref{fig:app:time_to_generate_tokens_llama_cpp:cpu} CPU and \subref{fig:app:time_to_generate_tokens_llama_cpp:gpu} A$100$ GPU. Measurements were carried out via the \texttt{llama-bench} tooling from \texttt{llama.cpp}. }
    \label{fig:app:time_to_generate_tokens_llama_cpp}
\end{figure}

Deriving exact measurements of how long is spent in attention layers in optimised inference libraries such as \texttt{llama.cpp} and \texttt{vLLM} can be non-trivial, due to limited existing tooling in their implementations and (to a lesser extent) probing models during inference may impact performance. \texttt{llama.cpp} features a benchmarking tool called \texttt{llama-bench}, which measures the time it takes to either prefill a model with a prompt of certain length, or autoregressively generate a sequence.

We employ \texttt{llama-bench}'s existing functionality to calculate the approximate time that is spent in attention, by observing that when generating a single token, the compute and data transfer associated with the attention layers scale linearly with the sequence length, with all other costs remaining constant. This can be seen in \cref{fig:app:time_to_generate_tokens_llama_cpp}, which shows the measured time it takes to generate a single token, given an existing sequence length $S$. As $S$ increases, the time it takes to generate a single token scales linearly.

From the measured benchmarks, lines of best fit were computed over the interquartile range of each $S$ (to reduce variance) for each hardware platform, which were found to be
\begin{equation}
    y_\text{\tiny CPU}(S) = (1.62 \times 10^{-5})S + 0.2438
\end{equation}
and
\begin{equation}
    y_\text{\tiny GPU}(S) = (4.686e \times 10^{-7})S + 0.009481
\end{equation}
for CPU and GPU respectively. The value of $y_\text{\tiny XPU}(0)$ corresponds to the time it takes for all non-attention data transfer and operations. Therefore, the proportion of time spent in attention, $z_\text{\tiny XPU}(S)$, can be approximated as
\begin{equation}
    z_\text{\tiny XPU}(S) \approx \frac{y_\text{\tiny XPU}(S) - y_\text{\tiny XPU}(0)}{y_\text{\tiny XPU}(S)}\text{,}
\end{equation}
the results of which can be seen in \cref{fig:llama_cpp_time_in_attention}.

\section{Attention Sparsity Analysis}
\label{sec:app:analysis}

\renewcommand\thefigure{\ref{sec:app:analysis}\arabic{figure}}
\setcounter{figure}{0}
\renewcommand\thetable{\ref{sec:app:analysis}\arabic{table}}
\setcounter{table}{0}
\renewcommand\theequation{\ref{sec:app:analysis}\arabic{equation}}
\setcounter{equation}{0}

\begin{figure}[h]
    \centering
    \subfloat[]{
    \includegraphics[width=7cm]{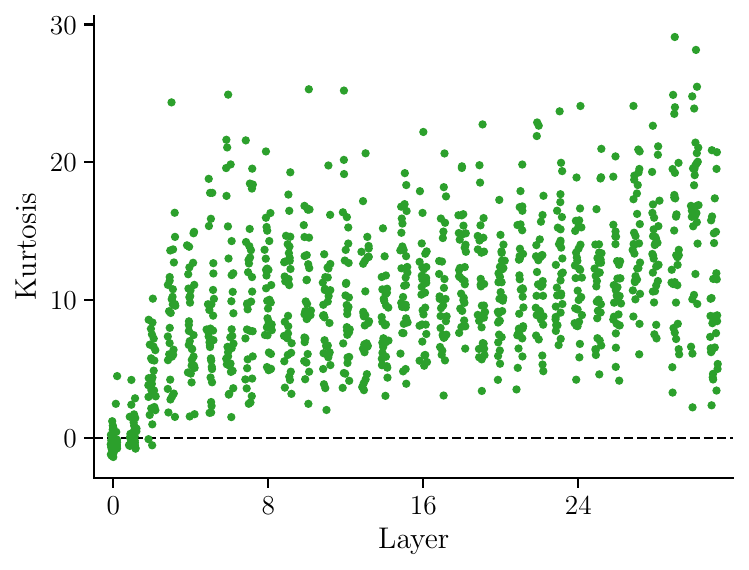}
    }
    \qquad
    \subfloat[]{
    \includegraphics[width=7cm]{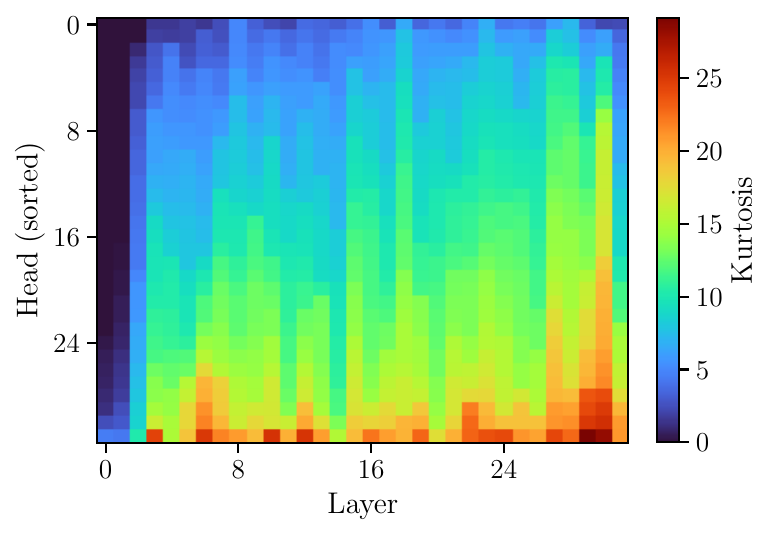}
    }
    \qquad
    \subfloat[]{
    \includegraphics[width=7cm]{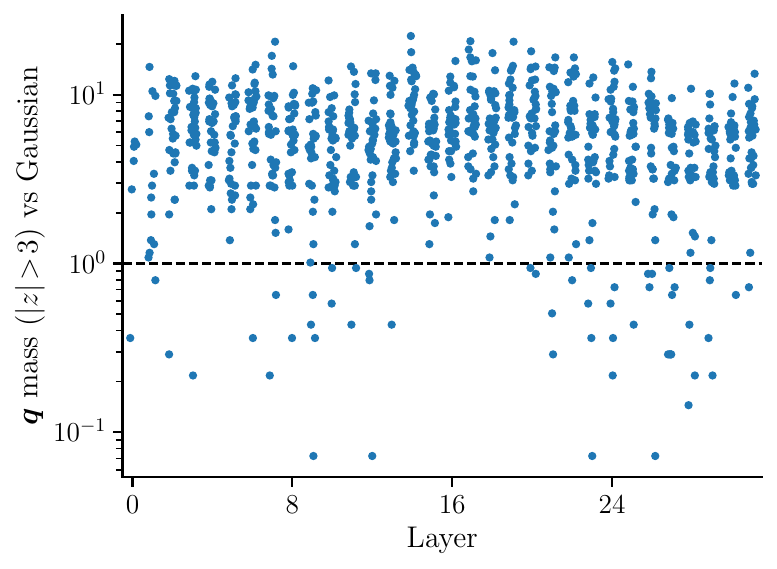}
    }
    \qquad
    \subfloat[]{
    \includegraphics[width=7.2cm]{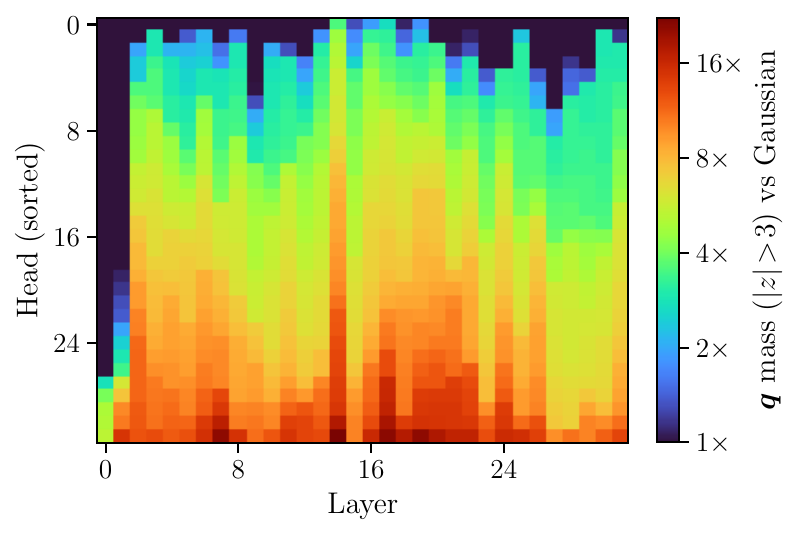}
    }
    \caption{Statistics of components of $\bm{q}$ for each head, as a function of layer. (Top) Kurtosis (Fisher), indicating that most heads have heavy-tailed $\bm{q}$. (Bottom) $z$-value mass, normalised by that of a Gaussian ($0.3\%$), showing that most heads are outlier-heavy. All Llama $2$ $7$B, measured over $40$ SQuAD examples.}
    \label{fig:app:analysis_extra_components}
\end{figure}

\begin{figure}[H]
    \centering
    \includegraphics[width=0.45\textwidth]{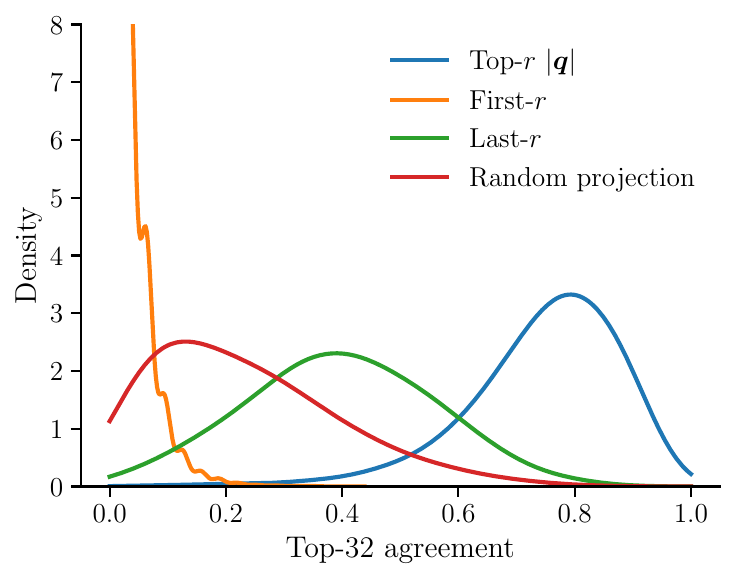}
    \caption{Top-$k$ agreement between approximate and true scores (Llama $2$ $7$B, measured over $40$ SQuAD examples). Top-$k$ agreement is the proportion of the top-$k$ positions that are correctly predicted by an approximated softmax, using a projection of $\boldsymbol{q}$, either component-wise or a random low-rank projection.}
    \label{fig:app:analysis_extra_agreement}
\end{figure}

In order to understand how to approximate attention in pretrained transformers, we analysed the queries, values and intermediate \textit{scores} vector (softmax output). We took $40$ examples from our SQuAD $1$-shot task, and generated the first completion token using the dense Llama $2$ $7$B model, capturing the $\boldsymbol{q}$ vector and $\boldsymbol{K}$, $\boldsymbol{V}$ matrices from every layer and attention head, showing derived statistics in \cref{fig:approximation_analysis,fig:app:analysis_extra_components,fig:app:analysis_extra_agreement}.

In \cref{fig:approximation_analysis:query_hist,fig:approximation_analysis:query_kurtosis_strip} we show that elements of the query vectors are not normally distributed, but have high sample kurtosis values. If compared to a normal distribution, the combined mass of the elements with absolute \textit{z-score} exceeding $3.0$ is up to $20\times$ higher. This leads us to theorise that query vectors in a pre-trained model inherently encode information \textbf{sparsely} using the tails. Therefore, the magnitude based sparsity we induce in the first stage of the algorithm does not significantly harm the approximation of the attention mappings.

We validate this claim by comparing the correspondence between the exact and approximated attention scores. \method{} uses the approximate attention scores to only choose the tokens that are important for the next generation step. The actual values of the approximate scores are not relevant, as these scores are not multiplied with value vectors and thus the property of interest to us is whether the top-$k$ indices in the approximate scores match those of the exact counterpart. This can be measured on a scale from $0$ to $1$, where $1$ means top-$k$ indices are identical between the approximation and the exact scores and $0$ means these sets do not overlap. We call this measure \emph{top-$k$ correspondence}. \Cref{fig:app:analysis_extra_agreement} provides an overview how the choice of rank and $k$ affects the top-$k$ correspondence aggregated over all attention heads of the model. We see that the query vector sparsity of $50\%$ and $75\%$ maintain high top-$k$ correspondence to the exact attention scores, which is consistently maintained over various values of $k$. Further analysis and a more detailed examination of top-$k$ correspondence is presented in \cref{sec:app:analysis}.

It is useful to drop positions in $\boldsymbol{V}$ given attention scores, but this can save at most half of the data transfers, since the whole of $\boldsymbol{K}$ is needed to calculate these scores. We propose approximating these scores using a subset of the components of $\boldsymbol{K}$. To test such an approximation, we measure the proportion of \textit{overlap} between the top $32$ positions in the approximated and true scores. If overlap is high, we can use the approximation to avoid transferring the whole $\boldsymbol{K}$ matrix, instead only transferring some components of $\boldsymbol{K}$ for all positions, then all components of $\boldsymbol{K}$ for some positions.

Our hypothesis is that the $r$ largest-magnitude components of $\boldsymbol{q}$ are most useful to predicting the score, $\boldsymbol{q} \boldsymbol{K}^\top$. The coverage of this technique against an arbitrary-component baseline is shown in \cref{fig:app:analysis_extra_agreement}. These results show that it is possible to achieve reasonably high overlap even using $r = d_h/8$, but that some later layers are harder to predict. Using the top-$r$ components outperforms the first $r$ baseline considerably.

\section{Benchmarking Detail}
\label{sec:app:benchmarking}

\renewcommand\thefigure{\ref{sec:app:benchmarking}\arabic{figure}}
\setcounter{figure}{0}
\renewcommand\thetable{\ref{sec:app:benchmarking}\arabic{table}}
\setcounter{table}{0}
\renewcommand\theequation{\ref{sec:app:benchmarking}\arabic{equation}}
\setcounter{equation}{0}

Benchmarking code is made available from:
\newline\url{https://github.com/graphcore-research/llm-inference-research/tree/2024-05-sparq}.

\paragraph{IPU measurements} We tested custom fully-fused Poplar implementations of both dense attention and \method{}, compiled using Poplar SDK 3.3.0+1403. On initialisation, we fill large $\boldsymbol{K}$ and $\boldsymbol{V}$ tensors with values $\sim N(0, 1)$ in streaming memory. On each benchmarking (outer) iteration, we first randomise the contents of a $\boldsymbol{q}$ in local memory, then perform multiple inner repeats of the attention op being profiled. We use $4$ inner repeats for dense attention, otherwise $1024/\mathrm{batch\_size}$, chosen because dense attention is much slower, and we swept a wide range of settings. We ran an outer loop of $2$ warm-up iterations followed by $10$ timed iterations, reporting the mean and standard error. The sweep covered $S \in [1024, 2048, \ldots, 65536]$, $\mathrm{batch\_size} \in [1, 4, 16, 64]$, \method{} $r \in [16, 32, 64]$ and $k \in [64, 128, 256, 512]$.

\paragraph{GPU measurements} All experiments use PyTorch 2.1.2+cu121 on Ubuntu AWS instances. To set up the experiment, we initialise the large $\boldsymbol{K}$ and $\boldsymbol{V}$ tensors with values $\sim N(0, 1)$. On each step, we draw $\boldsymbol{q} \sim N(0, 1)$, run \texttt{torch.cuda.synchronize} before starting a host-side wall-clock timer, run the op, and synchronize again before stopping the timer. We run $20$ warm-up iterations followed by $200$ timed iterations, reporting mean and standard error. For dense baseline implementations, we tested a vanilla PyTorch implementation, with/without \texttt{torch.compile} and \texttt{torch.nn.functional.scaled\_dot\_product\_attention}, selecting each backend (math, flash, mem\_efficient) manually. For \method{} implementations, we tested vanilla PyTorch (lightly hand-optimised from \cref{sec:app:code}), with/without \texttt{torch.compile}. We also toggled fused gather-matmul kernels written in Triton, and whether $\boldsymbol{K}$ was stored twice in $S$-contiguous (for \textbf{Step 1}) and $d_h$-contiguous (for \textbf{Step 2}) layouts, or only once in $d_h$-contiguous layout. We tested $S \in [1024, 2048, 4096, 8192, 16384]$, $\mathrm{batch\_size} \in [1, 4, 16, 64]$, \method{} $r \in [16, 32, 64]$ and $k \in [64, 128, 256, 512]$.

\begin{figure}
    \centering
    \includegraphics[width=10cm]{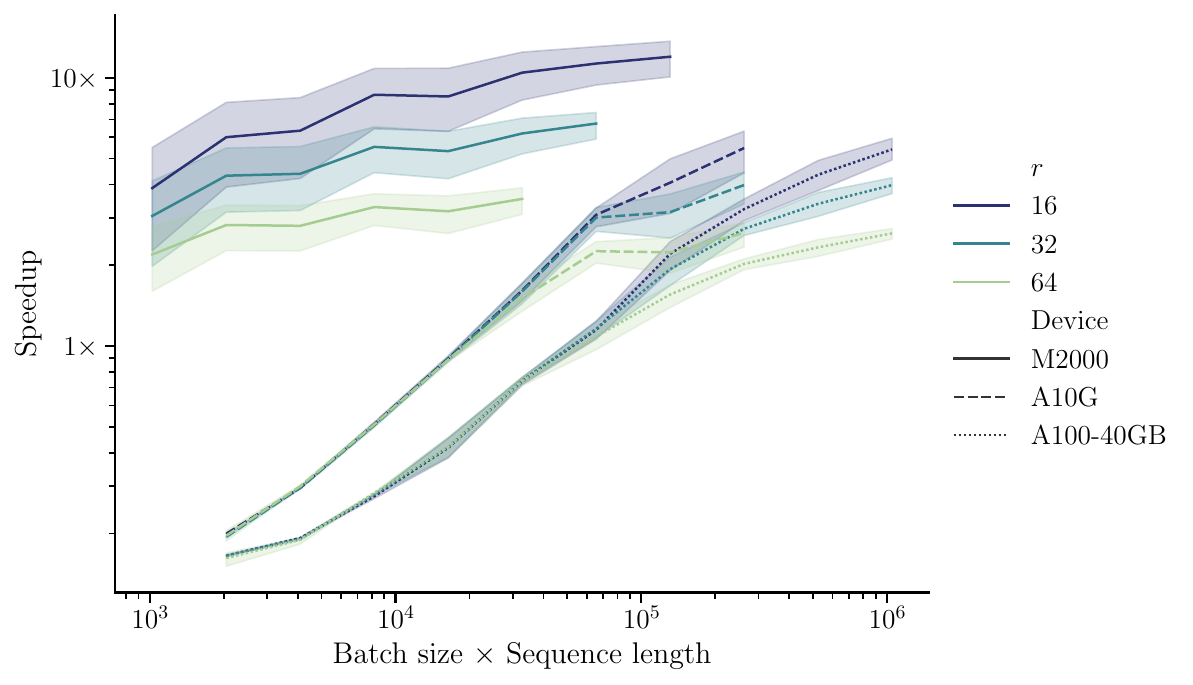}
    \caption{SparQ speedup over the dense baseline, across a range of batch size ($1$-$64$), sequence length ($1024$-$65536$) and $k$ ($64$-$512$), for different devices. We note that for both GPUs, the number of KV elements is a limiting factor for the achieved speedup, and that this could be improved by writing a fully fused SparQ Attention kernel.}
    \label{fig:app:benchmark_scaling}
\end{figure}

\paragraph{Additional results} In addition to the headline results shared in \cref{sec:benchmarking} and \cref{fig:benchmark_a100}, we give an aggregate picture of the trends in \cref{fig:app:benchmark_scaling}. Since the number and dimension of heads is fixed, the x-axis is proportional to the size of the input tensors. On IPU (M$2000$), strong speedups are available across a range of input sizes, principally depending on $r$, but also on $k$ (not shown). On GPU, sufficient input size is required to observe a speedup over the dense baseline, with the more bandwidth-limited A10G reaching speedups sooner. While part of this effect can be linked to the fundamental additional complexity of \method{}, we anticipate that small input sizes could be accelerated considerably with additional kernel fusion. With an appropriate limit to sequence length, \method{} could even be fused into a single CUDA kernel.

\paragraph{Storing $\bm{K}$ twice} One limitation of a theoretical model of data transfer is that it does not account for the granularity of memory access. Since the $\boldsymbol{K}$ matrix is indexed on different axes in \textbf{Step 1} and \textbf{Step 2} of \method{}, a naive implementation would fetch non-contiguous elements in one of the two steps. To mitigate this, we propose storing $\boldsymbol{K}$ twice, once in $S$-major format and once in $d_h$-major format. This increases KV cache memory usage by $50\%$, but uses only a small amount of extra bandwidth to write $\boldsymbol{k}$ twice. This extra write is non-contiguous, but small, so should not form a bottleneck.

\section{Methodology}
\label{sec:app:method}

\renewcommand\thefigure{\ref{sec:app:method}\arabic{figure}}
\setcounter{figure}{0}
\renewcommand\thetable{\ref{sec:app:method}\arabic{table}}
\setcounter{table}{0}
\renewcommand\theequation{\ref{sec:app:method}\arabic{equation}}
\setcounter{equation}{0}

We provide a comprehensive description of our experimental setup for reference in \cref{app:table:hyperparameters}.

\paragraph{Baselines}
We use our own implementation of H$_2$O \citep{zhang2023h2o}, which differs from the authors' implementation in that it uses a fixed cache size $k$, rather than a fixed proportion of the current sequence length. To validate that these implementations are sufficiently similar, we ran their implementation through our harness on a small model and sample size. On SQuAD $1$-shot, with Pythia-$1.4$B, using $k=256$, $l=64$, our implementation was correct for $60$ of $200$ examples, theirs for $57$ (the dense baseline achieved $74$). Perhaps more importantly, we found that of the $79$ times that either output differed from dense, $41$ occurrences showed a $20$-character prefix match between our implementation and theirs. The fact that the two implementations often generate the same errors (despite minor implementation differences) reassures us that our results should be a fair representation of H$_2$O.

\paragraph{Compression ratio} We define the compression ratio as the ratio of attention data transfers required for the sparse technique and the dense data transfers. Similarly as we have derived the transfers for \method{} in \cref{sec:sparseq}, we can define the transfers required for each baseline technique:
\begin{align*}
    \mathcal{M}_{\mathrm{H_2O}} &= 2\,k\,d_h + 2\,d_h + 2\,S\\
    \mathcal{M}_{\mathrm{LM_{\infty}}} &= 2\,k\,d_h + 2\,d_h\\
    \mathcal{M}_{\mathrm{FlexGen}} &= S\,d_h + k\,d_h + 2\,d_h
\end{align*}

and $\mathcal{M}_{\mathrm{dense}} = 2\,S\,d_h + 2\,d_h$. For each technique, the compression ratio is then $\mathcal{M}_{\mathrm{technique}}/ \mathcal{M}_{\mathrm{dense}}$.
\setcounter{table}{0}
\renewcommand\thetable{\ref{sec:app:method}\arabic{table}}

\newpage
\vspace*{\fill}
\begin{table}[H]
\centering
\renewcommand{\arraystretch}{1.25}

\begin{tabular}{ | c | r | p{6.4cm} | }
\hline
 
 \multirow{8}{*}{\textbf{Models}} & \multirow{2}{*}{Llama $2$} & $7$B ($d_h = 128$, Max $S = 4096$, $g = 1$)\\
 & & $13$B ($d_h = 128$, Max $S = 4096$, $g = 1$)\\\cline{2-3}
 & Llama 3 & $8$B ($d_h = 128$, Max $S = 8192$, $g = 4$)\\\cline{2-3}
 & Mistral & $7$B ($d_h = 128$, Max $S = 8192$, $g = 4$)\\\cline{2-3}
 & Gemma & $7$B ($d_h = 256$, Max $S = 8192$, $g = 1$)\\\cline{2-3}
 & \multirow{3}{*}{Pythia} & $1.4$B ($d_h = 128$, Max $S = 2048$, $g = 1$)\\
 & & $2.8$B ($d_h = 80$, Max $S = 2048$, $g = 1$)\\
 & & $6.9$B ($d_h = 128$, Max $S = 2048$, $g = 1$)\\\hline

 \multirow{5}{*}{\textbf{Tasks}}
 & \multirow{2}{*}{Question Answering} & SQuAD $1$-shot ($4000$ samples)\newline TriviaQA $0$-shot ($2992$ samples) \\\cline{2-3}
 & Summarisation & CNN/DailyMail $0$-shot ($500$ samples) \\\cline{2-3}
 & Language Modelling & WikiText-$103$ LM ($500$ samples) \\\cline{2-3}
 & Artificial & Repetition ($1000$ samples) \\\hline

 \multirow{7}{*}{\textbf{Baselines}}
 & \multirow{3}{*}{H$_2$O} & keep $(k-l)$ tokens with highest $\mathrm{score}(n) = \sum_{i}s_{in}$ and the most recent $l = k/4$
    \newline $k \in \{192, 256, 384, 512, 768\}$
    \\\cline{2-3}

 & \multirow{2}{*}{LM-Infinite} & take the first $16$ tokens, and most recent $k-16$
    \newline $k \in \{192, 256, 384, 512, 768\}$ \\\cline{2-3}
    
 & \multirow{2}{*}{FlexGen} & take top-$k$ tokens using exact attention scores \newline $k \in \{ 2, 8, 32, 128, 256 \}$ \\\hline
    
 \multirow{3}{*}{\textbf{\method}}
 & Rank $r$ & $\{8, 16, 32, 64\}$ \\\cline{2-3}
 & Number of values $k$ & $128$ \\\cline{2-3}
 & Local window $l$ & $k/4$ \\\hline

\end{tabular}
\vspace{0.5cm}
\caption{Experimental setup.}
\label{app:table:hyperparameters}
\end{table}

\renewcommand{\thetable}{\ref{sec:app:method}2}


\vspace*{\fill}
\newpage
\subsection{Examples}

We illustrate the task setup with a single example per task, showing the prompt formatting and a cherry-picked example. In each case, we show outputs from a dense Llama $2$ $13$B model, \method{} ($r=8, k=128$), H$_2$O and LM-Infinite ($k=192$). Where ``\texttt{...}'' appears, we have truncated the line of text for brevity.

\subsubsection{Question Answering (SQuAD $1$-shot)}

\begin{small}
\begin{verbatim}### PROMPT (5725e1c4271a42140099d2d9)
Title: University of Chicago. Background: Current ...
Title: Harvard University. Background: Harvard has...
Title: Oxygen. Background: In one experiment, Lavo...
Title: Oxygen. Background: Oxygen storage methods ...
Title: Fresno, California. Background: This vibran...
Title: Fresno, California. Background: Before Worl...
Title: Steam engine. Background: The working fluid...
Title: Sky (United Kingdom). Background: While BSk...
From what you've just read about Fresno, California, please answer the following questions.
Question: Where is Audra McDonald from?
Answer: Fresno
Question: In what year did Roger Rocka's Dinner Theater & Good Company Players open?
Answer:

### OUTPUT
      DENSE: 1978
      SPARQ: 1978
        H2O: 1979
LM-INFINITE: 1975 (Roger Rock
\end{verbatim}
\end{small}

\subsubsection{Question Answering (TriviaQA 0-shot)}

\begin{small}
\begin{verbatim}### PROMPT (dpql_5685)
Apéritifs and digestifs ( and) are drinks, typical...
Apéritifs
An apéritif is an alcoholic beverage usually serve...
"Apéritif" may also refer to a snack that precedes...
"Apéritif" is a French word derived from the Latin...
...
...
* Distilled liquors (ouzo, tequila, whisky or akva...
* Liquor cocktails (Black Russian, Rusty Nail, etc...
In certain areas, it is not uncommon for a digesti...
Bitter digestifs typically contain carminative her...
In many countries, people drink alcoholic beverage...
Question: Which aperitif is named for the Paris chemist who created it in 1846?
Answer:

### OUTPUT
      DENSE: Dubonnet
      SPARQ: Dubonnet
        H2O: Dubonnet
LM-INFINITE: Byrrh
\end{verbatim}
\end{small}

Note that for Pythia, the prompt ``Single-word answer:'' was used in place of ``Answer:'', as this helped prevent the model from restating the question in the answer (often qualitatively correct, but not a regex match).

\newpage
\subsubsection{Summarisation (CNN/DailyMail)}

\begin{small}
\begin{verbatim}### PROMPT (a62bbf503be06e8b1f8baa4f3cd537310d5aa3bc)
Article: Prince William arrived in China tonight for one of the most high-profil...
Summary:

### OUTPUT
      DENSE: Prince William arrived in China tonight for one of the most high-profile – ...
      SPARQ: Prince William arrived in China tonight for one of the most high-profile – ...
        H2O: Prince William arrived in China tonight for a high-profile visit that will ...
LM-INFINITE: Prince William and Kate Middleton are in Japan for a three-day tour. The ro...
\end{verbatim}
\end{small}

\subsubsection{Repetition (Shakespeare)}

\begin{small}
\begin{verbatim}### PROMPT (210496)
 you mistake me much;
I do lament the sickness of the king.
...
...
Peace, children, peace! the king doth love you well:
Incapable and shallow innocents,
You cannot guess who caused your father's death.

Boy:
Grandam, we can; for my good uncle Gloucester
Told me, the king, provoked by the queen,
Devised impeachments to imprison him :
And when my uncle told me so, he wept,
And hugg'd me in his arm, and kindly kiss'd my cheek;
...
...
 the king doth love you well:
Incapable and shallow innocents,
You cannot guess who caused your father's death.

Boy:
Grandam, we

### OUTPUT
      DENSE: can; for my good uncle Gloucester
      SPARQ: can; for my good uncle Gloucester
        H2O: can;
LM-INFINITE: 'll not stand to prate, but to the purpose.
\end{verbatim}
\end{small}

\newpage
\subsubsection{Language Modelling (WikiText-103)}

\begin{small}
{\fontfamily{cmtt}\selectfont
\begin{verbatim}### QUERY (2)

 = Mellor hill fort = 
 
 Mellor hill fort is a prehistoric site in North West England , that dates from ...
 
 = = Location = = 
 
 Mellor lies on the western edge of the Peak District in the Metropolitan Boroug...
 
 = = Background = = 
 
 Until the 19th century little was known about hill forts ; none had been excava...
 The study of hill forts was popular in the 19th century , with a revival in the...
 
 = = History = = 
 
 There is evidence of human activity on the site pre @-@ dating the Iron Age , a...
 A flint dagger was discovered on the site . This type of artefact is rare in Gr...
 The hill fort was built in and used throughout the Iron Age , as demonstrated b...
\end{verbatim}
\ Fragments of glass , possibly Roman in origin , and shards of pottery which date to the 1st and 2nd centuries AD , indicate the site was used in the Romano @-@ British period . However{\color{blue}\ no Roman structures have been discovered , and the nature of Roman activity at the site is a source of speculation . The position of the hilltop indicate that it was easily defended ; however , local finds indicate it was a high @-@ status settlement rather than a military outpost unless a similar feature was located nearby . One reason that Roman structures have not been identified is that the Romano}
\begin{verbatim}### BPC
      DENSE: 0.669
      SPARQ: 0.673
        H2O: 0.685
LM-INFINITE: 0.692
\end{verbatim}
}\end{small}


\end{document}